\documentclass[twoside,11pt]{article}
\usepackage{jmlre}

\usepackage{amsfonts}
\usepackage{amssymb}
\usepackage{amsmath}
\usepackage{enumerate,ifthen,epic,fancybox,amsthm,fancyhdr}
\usepackage{natbib}
\usepackage{verbatim}
\usepackage{rotating}
\usepackage[table]{xcolor}
\usepackage{booktabs}
\usepackage{caption}
\usepackage{lscape}
\usepackage{mathrsfs}
\usepackage[export]{adjustbox}
\usepackage{subfig}
\usepackage{enumerate}
\usepackage{url}
\usepackage{here}
\usepackage{commath}
\usepackage{multirow}
\usepackage{gensymb}
\usepackage{epsfig}
\usepackage{float}
\usepackage{fontenc}

\DeclareMathOperator{\Tr}{Tr}
\def\diag{{\rm diag}}

\ShortHeadings{Fast Automatic Smoothing for Generalized Additive Models}{Yousra El-Bachir and Anthony Davison}
\firstpageno{1}

\begin{document}

\title{Fast Automatic Smoothing for Generalized Additive Models}
\thispagestyle{plain}
\author{\name \centering Yousra El-Bachir \quad yousra.elbachir@gmail.com\\
	     Anthony~C.~Davison \quad anthony.davison@epfl.ch \\
	     \medskip
	    \addr \centering EPFL-FSB-MATH-STAT\\
	    \addr Ecole Polytechnique F\'ed\'erale de Lausanne\\
	     Station 8, Lausanne 1015, Switzerland \\
	     \medskip
	    \medskip
	    \medskip 
	    }

\maketitle

\begin{abstract}
Multiple generalized additive models (GAMs) are a type of distributional regression wherein parameters of probability distributions depend on predictors through smooth functions, with selection of the degree of smoothness via $L_2$ regularization. Multiple GAMs allow finer statistical inference by incorporating explanatory information in any or all of the parameters of the distribution. Owing to their nonlinearity, flexibility and interpretability, GAMs are widely used, but reliable and fast methods for automatic smoothing in large datasets are still lacking, despite recent advances. We develop a general methodology for automatically learning the optimal degree of $L_2$ regularization for multiple GAMs using an empirical Bayes approach. The smooth functions are penalized by different amounts, which are learned simultaneously by maximization of a marginal likelihood through an approximate expectation-maximization algorithm that involves a double Laplace approximation at the E-step, and leads to an efficient M-step. Empirical analysis shows that the resulting algorithm is numerically stable, faster than all existing methods and achieves state-of-the-art accuracy. For illustration, we apply it to an important and challenging problem in the analysis of extremal data. 
\end{abstract}



\section{Introduction}\label{sec:litrev}
Generalized additive models (GAMs) are supervised learning tools that describe the relationship between response variables and predictors using additive smooth functions~\citep{hastie1986}.  These were originally represented by scatterplot smoothers and trained by backfitting~\citep{BreimanFriedman},  implemented in the \verb|R| \citep{RR} package \verb|gam| that stems from \citet{HastieTib}, which selects the level of smoothness by stepwise regression using approximate distributional results. Backfitting allows smooth terms to be represented by local regression smoothers \citep{Loess1993}, but inference based on the resulting fit is awkward. \citet{YeeWild1996} later proposed modified vector backfitting, whereby several smooth responses are learned simultaneously. Their method, embodied in the package \verb|VGAM|, first learns the linear components and then learns the nonlinear part by training a vector additive model on the resulting partial residuals. In the package \verb|gamlss|, \cite{RigbyStasinopoulos2005} learn the smooth functions sequentially by combining backfitting with two separate algorithms, which optimize the penalized likelihood of the regression weights. The first algorithm generalizes that of \citet{ColesGreen1992}, whereas the second generalizes that of \citet{Rigby1996a}, and is preferable when the parameters of the distribution are orthogonal with respect to the information matrix. All these approaches invoke backfitting, which dissociates learning of the regression model from that of the smoothing parameters. This may be statistically inefficient, and accuracy may be increased by learning the appropriate degree of smoothing as part of the regression training. 

An alternative representation of GAMs that enables automatic smoothing is via basis function expansion using reduced rank smoothing; this is the foundation upon which we build our methodology. We suppose that independent observations come from a probability distribution whose parameters are explained by generalized additive models. Let $Y_i$ denote a random variable with realized value $y_i$ and probability distribution function $F_i(y_i;\theta_i)$ that depends on a parameter vector $\theta_i=(\theta_i^{(1)}, \ldots, \theta_i^{(D)}) \in \mathbb{R}^D$; so for the training set $\boldsymbol{y}=(y_1, \ldots, y_n)^T$, the full parameter vector is $\boldsymbol{\theta} = (\theta_1, \ldots, \theta_n)^T\in \mathbb{R}^{nD}$ with subvectors $\boldsymbol{\theta}^{(d)}=(\theta_1^{(d)}, \ldots, \theta_n^{(d)})^T \in \mathbb{R}^n$ for $d=1, \ldots, D$. In the Gaussian model for example, $D=2$, $\boldsymbol{\theta}^{(1)}=\boldsymbol{\mu}$ is the mean and $\boldsymbol{\theta}^{(2)}=\boldsymbol{\sigma}$ is the standard deviation, and we have $\boldsymbol{\theta}=(\mu_1, \sigma_1, \ldots, \mu_n, \sigma_n)^T$. For a multiple generalized additive model, each $\boldsymbol{\theta}^{(d)}$ has an additive structure, which we now describe. Let ${X_i^{(d)}}^*$ denote the $i$-th row of a feature matrix corresponding to a parameter vector ${\beta^{(d)}}^*$ that includes an offset. Let $q_d \geqslant 0$ denote the number of unknown smooth functions $f_j^{(d)}$ contributing to $\boldsymbol{\theta}^{(d)}$, and let $x_s, x_t, \ldots$ denote the predictors. The components $\theta_i^{(d)}$ of $\boldsymbol{\theta}^{(d)}$ represent a GAM through 
\begin{eqnarray*}
\theta_i^{(d)} = {X_i^{{(d)}}}^{*}{\beta^{(d)}}^{*} + \sum_{j=1}^{q_{d}}f_j^{(d)}(x_{is},x_{it},\ldots), \quad i=1, \ldots, n,
\end{eqnarray*}
where each of the $f_j^{(d)}$ can be a function of one or more predictors, and is represented as an expansion of basis functions $b^{(d)}_k(x)$, splines for example, whose weights are the regression parameters
\begin{eqnarray*}
f^{(d)}_j(x) = \sum_{k=1}^K \beta_k b^{(d)}_k(x),
\end{eqnarray*}
where the basis dimension $K$ is chosen manually and typically grows slowly with the size $n$ of the training set. In this setting, the components of $\boldsymbol{\theta}^{(d)}$ become $\theta_i^{(d)}= X_i^{(d)}\boldsymbol{\beta}^{(d)}$, where $\boldsymbol{\beta}^{(d)} \in \mathbb{R}^{p_{d}}$ and $X^{(d)}  \in \mathbb{R}^{n\times p_{d}}$ denote respectively the regression weights and the feature matrix, including their parametric parts. We assume that the columns of $X^{(d)}$ have been transformed to absorb sum-to-zero identifiability constraints on the smooth functions. The smoothness of $f_j^{(d)}$ is adjusted by a quadratic penalty on its curvature
\begin{eqnarray*}
{\rm{PEN}}(\lambda_j^{(d)}) = \lambda_j^{(d)}\int \left\{{f_j^{(d)}}^{\prime \prime}(t)\right\}^2 \, \mathrm{d}t =\lambda_j^{(d)} {\boldsymbol{\beta}^{(d)}}^T S_j^{(d)} \boldsymbol{\beta}^{(d)} \in \mathbb{R}, 
\end{eqnarray*}
where the positive regularization parameter $\lambda_j^{(d)}$ controls the degree of smoothness and $S^{(d)}_j \in \mathbb{R}^{p_{d}\times p_{d}}$ is a known symmetric and semi-positive definite smoothing matrix. 
On defining analogous quantities for any of the parameter vectors $\boldsymbol{\theta}^{(1)},\ldots, \boldsymbol{\theta}^{(D)}$, and stacking together the regression weights and the smoothing parameters to form $\boldsymbol{\beta}\in \mathbb{R}^p$ and $\boldsymbol{\lambda} \in \mathbb{R}^q$ with $p=\sum_{d=1}^Dp_{d}$ and $q=\sum_{d=1}^Dq_{d}$, the full weight vector and curvature penalties are parametrized by
\begin{eqnarray}
\boldsymbol{\theta}_{\boldsymbol{\beta}}= \boldsymbol{X}\boldsymbol{\beta} \in \mathbb{R}^{nD}, \quad {\rm{PEN}}(\boldsymbol{\lambda}) = \sum_{d=1}^D\sum_{j=1}^{q_d} {\rm{PEN}}(\lambda_j^{(d)})=  \boldsymbol{\beta}^T \boldsymbol{S}_{\boldsymbol{\lambda}} \boldsymbol{\beta} \in \mathbb{R}, \label{eq:Penalty} 
\end{eqnarray}
where the $i$-th row block of the full feature matrix $\boldsymbol{X}\in \mathbb{R}^{nD \times p}$ is 
\begin{eqnarray*}
\boldsymbol{X}_i = \diag \left(X_i^{(1)},\ldots, X_i^{(D)}\right) \in \mathbb{R}^{D\times p},
\end{eqnarray*} and the full smoothing matrix 
\begin{eqnarray}
\boldsymbol{S}_{\boldsymbol{\lambda}}=\diag \left(\lambda_1^{(1)}S_1^{(1)}, \ldots, \lambda_{q_D}^{(D)}S_{q_D}^{(D)}\right) \in \mathbb{R}^{p\times p} \label{eq:SblockDiag}
\end{eqnarray} 
is block diagonal.

Learning the regression weights involves balancing the conflicting goals of providing a good fit to the data and avoiding overfitting. For a given $\boldsymbol{\lambda}$, this is obtained by maximizing the penalized log-likelihood for $\boldsymbol{\beta}$,
\begin{eqnarray}
\ell_{\rm P}(\boldsymbol{\beta}; \boldsymbol{y}, \boldsymbol{\lambda}) = \ell_{\rm L}\left(\boldsymbol{\theta}_{\boldsymbol{\beta}}; \boldsymbol{y}\right) - \cfrac{1}{2}\ \boldsymbol{\beta}^T\boldsymbol{S}_{\boldsymbol{\lambda}}\boldsymbol{\beta}, \label{eq:lp}
\end{eqnarray} 
where the log-likelihood $\ell_{\rm L}$ may be written equivalently in terms of $\boldsymbol{\theta}$ or of $\boldsymbol{\beta}$. With $U(\boldsymbol{\beta})\in  \mathbb{R}^{p}$ and $H(\boldsymbol{\beta})\in  \mathbb{R}^{p\times p}$, or $U(\boldsymbol{\theta})\in  \mathbb{R}^{Dn}$ and $H(\boldsymbol{\theta})\in  \mathbb{R}^{Dn\times Dn}$,  the gradient and negative Hessian of $\ell_{\rm L}$ with respect to $\boldsymbol{\beta
}$ and to $\boldsymbol{\theta}$, the corresponding penalized quantities are 
\begin{align}
U_{\rm P}(\boldsymbol{\beta}; \boldsymbol{\lambda}) &=U(\boldsymbol{\beta})-\boldsymbol{S}_{\boldsymbol{\lambda}}\boldsymbol{\beta},           &  H_{\rm P}(\boldsymbol{\beta}; \boldsymbol{\lambda}) &= H(\boldsymbol{\beta})+\boldsymbol{S}_{\boldsymbol{\lambda}},\label{eq:Hp} \\
&=\boldsymbol{X}^T U(\boldsymbol{\theta})-\boldsymbol{S}_{\boldsymbol{\lambda}}\boldsymbol{\beta},         &  &=\boldsymbol{X}^T H(\boldsymbol{\theta})\boldsymbol{X}+\boldsymbol{S}_{\boldsymbol{\lambda}} \nonumber.
\end{align}
The negative Hessian is used for calculating standard errors and confidence intervals. Maximization of the penalized log-likelihood \eqref{eq:lp} provides an estimator for $\boldsymbol{\beta}$ for a given value of the smoothing parameters $\boldsymbol{\lambda}$. We now review the main frequentist methods for embodying learning of $\boldsymbol{\lambda}$ in that of the regression weights. The two strategies for this optimize a criterion for the smoothing parameters whilst updating the regression weights: performance iteration \citep{Gu1992}, and outer iteration \citep{OSullivan1986}. In the first, the updating step consists of one iteration for the smoothing parameters, followed by one iteration for the regression weights---often performed by iterative weighted least squares~\citep{nelderwedderburn1972}. Since a new trial for the smoothing parameters does not require the convergence of the regression model, performance iteration is computationally efficient if it converges, but as the smoothness selection criterion changes from iteration to iteration with the intermediate estimate of the regression model, convergence is not guaranteed; indeed, \citet{Wood2008,Wood2011} shows that this strategy can fail. Outer iteration comprises one update for the smoothing parameters followed by one full optimization for the regression weights. Since the former are obtained from a regression model that is fixed from iteration to iteration, the convergence of outer iteration can be guaranteed, but each updating step is computationally more expensive, and the dependence between the regression weights and the smoothing parameters is more challenging to elucidate.

The strategy for automatic smoothing being set, the classical approach for choosing its tuning parameters is to minimize measures of prediction error such as the Akaike or Bayesian information criteria, AIC or BIC, or the generalized cross-validation (GCV) criterion. The first tends to overfit, BIC presupposes that one of the learned models is correct, and GCV can generate multiple minima and unstable estimates that may lead to substantial underfitting~ \citep{ReissOgden,Wood2008}. Use of marginal likelihood overcomes these limitations, but involves intractable integrals. Despite the wide use of GAMs, automatic learning of their smoothing parameters is still an open problem. The reliable method~\citep{Wood2011} and its generalization~\citep{Wood2016}, implemented in the \texttt{R} recommended package \texttt{mgcv}, combine the advantages of the marginal likelihood approach with the good convergence of outer iteration. However, they are challenging to set up, difficult to extend to new families of distributions, and are computationally expensive for large datasets. On the other hand, methods specifically designed for large~\citep{Wood2015} and big~\citep{WoodBig} datasets are based on performance iteration, and so offer no guarantee of convergence. In this paper we overcome these limitations by presenting a new approach that is simpler, faster and achieves state-of-the-art accuracy. 

The rest of the paper is organized as follows. Section~\ref{sec:SmoothingParam} introduces our proposed automatic smoothness selection procedure, which is based on an approximate expectation-maximization algorithm. Section~\ref{sec:simulation} assesses its performance with a simulation study. Section~\ref{sec:dataAnalys} provides a real data analysis on extreme temperatures, and Section~\ref{sec:discussion} closes the paper with a discussion.


\section{Automatic smoothing}\label{sec:SmoothingParam}
The Bayesian formalism provides an interpretation for the smoothing penalty that underlies the weighted $L_2$ regularization in~\eqref{eq:Penalty}, as we now describe. Let $\boldsymbol{S}_{\boldsymbol{\lambda}}^{-}$ denote the generalized inverse of $\boldsymbol{S}_{\boldsymbol{\lambda}}$, and suppose that the regression weights have an improper multivariate Gaussian prior density $\mathcal{N}(0,\boldsymbol{S}_{\boldsymbol{\lambda}}^{-})$ \citep{kimeldorf1970,Silverman1985}
\begin{eqnarray}
\pi(\boldsymbol{\beta}; \boldsymbol{\lambda})=\left(2\pi\right)^{-(p-m)/2} \ \left| \boldsymbol{S}_{\boldsymbol{\lambda}} \right|^{1/2}_+  \exp{\left(-\frac{1}{2} \boldsymbol{\beta}^T\boldsymbol{S_{\lambda}}\boldsymbol{\beta} \right)}, \label{eq:prior}
\end{eqnarray}
where $m$ is the number of zero eigenvalues of $\boldsymbol{S}_{\boldsymbol{\lambda}}$ and $|\boldsymbol{S}_{\boldsymbol{\lambda}}|_+$ is the product of its positive eigenvalues. With $f$ denoting the density of the data, the log-posterior density for $\boldsymbol{\beta}$ is
\begin{eqnarray}
\ell(\boldsymbol{\beta} \mid \boldsymbol{y}; \boldsymbol{\lambda}) &=& \log\left\{ f(\boldsymbol{y} \mid \boldsymbol{\beta}; \boldsymbol{\lambda})\ \pi(\boldsymbol{\beta}; \boldsymbol{\lambda})\right\} - \log f(\boldsymbol{y}; \boldsymbol{\lambda}) \label{eq:margconst} \\
&=& \ell_{\rm P}(\boldsymbol{\beta}; \boldsymbol{y}, \boldsymbol{\lambda}) + \cfrac{1}{2} \log \left| \boldsymbol{S}_{\boldsymbol{\lambda}} \right|_+  -\dfrac{p-m}{2} \log\left(2\pi \right) - \log f(\boldsymbol{y}; \boldsymbol{\lambda}).\label{eq:posterior}
\end{eqnarray}
The smoothing penalty \eqref{eq:Penalty} now appears as the key component of the logarithm of the prior~\eqref{eq:prior}, and the penalized log-likelihood~\eqref{eq:lp} as the log-posterior~\eqref{eq:posterior} (up to a constant depending on $\boldsymbol{\lambda}$). The smoothing parameters can hence be learned from the last term on the right  of~\eqref{eq:margconst}, the marginal density of $\boldsymbol{y}$,
\begin{eqnarray*}
L_{\rm M}(\boldsymbol{\lambda};\boldsymbol{y}) =  f(\boldsymbol{y}; \boldsymbol{\lambda}) = \displaystyle \int f(\boldsymbol{y}, \boldsymbol{\beta}; \boldsymbol{\lambda}) \,\mathrm{d}\boldsymbol{\beta}
= \displaystyle \int f(\boldsymbol{y}\mid \boldsymbol{\beta}; \boldsymbol{\lambda})\ \pi(\boldsymbol{\beta}; \boldsymbol{\lambda}) \,\mathrm{d}\boldsymbol{\beta}.
\end{eqnarray*}
A fully Bayesian approach would involve choosing a prior density for $\boldsymbol{\lambda}$ and integrating out over it, but instead we take an empirical Bayes approach and transform the smoothness selection problem to an optimization problem, where the optimal $\boldsymbol{\lambda}$ are the maximizers of the log-marginal likelihood
\begin{eqnarray}
\ell_{\rm M}(\boldsymbol{\lambda};\boldsymbol{y}) &=& \log L_{\rm M}(\boldsymbol{\lambda};\boldsymbol{y}) \equiv \cfrac{1}{2}\log \left| \boldsymbol{S}_{\boldsymbol{\lambda}} \right|_+ + \log \displaystyle \int \exp \ell_{\rm P}(\boldsymbol{\beta}; \boldsymbol{y}, \boldsymbol{\lambda})  \, \mathrm{d}\boldsymbol{\beta}. \label{eq:explp}
\end{eqnarray}
The integral over $\boldsymbol{\beta}$ is intractable, and is typically approximated by importance sampling, quadrature or Laplace approximation. Importance sampling is a Monte Carlo integration technique under which the integral is treated as an expectation, but its performance relies on the choice of the distribution from which to sample, and its accuracy increases only with the number of samples. Quadrature involves a discretization of the integrand over the domain of integration, and amounts to calculating a weighted sum of the values of the integrand. Both methods perform well when the number of regression weights is small, but become computationally infeasible for $p>10$. The most common deterministic approach is Laplace approximation, which yields an analytical expression for~\eqref{eq:explp} by exploiting quadratic Taylor expansion of the log-integrand around the maximum penalized likelihood estimate. However, optimization of the resulting approximate log-marginal likelihood has several drawbacks. Each updating step includes intermediate maximizations, involves unstable terms that need careful and computationally expensive decompositions, and requires the fourth-order derivatives of the log-likelihood. These make Laplace approximation computationally demanding for smoothness selection, and limit its extension to complex models~\citep{Wood2011,Wood2016}. In this paper we present an alternative approach that is easier to implement, faster and achieves state-of-the-art accuracy. 

\subsection{Approximate expectation-maximization}
We directly maximize the log-marginal likelihood~\eqref{eq:explp} with respect to the smoothing parameters and circumvent evaluation of its approximation using the expectation-maximization (EM) algorithm~\citep{EM77,EMbook2008}. The EM algorithm is an iterative method for computing maximum likelihood estimators for difficult functions by alternating between an expectation step, the E-step, and its maximization, the M-step, at every iteration until convergence. Ignoring the constant term, taking conditional expectations of equation~\eqref{eq:posterior}  with respect to the posterior $\pi(\boldsymbol{\beta} \mid \boldsymbol{Y}=\boldsymbol{y}; \boldsymbol{\lambda}_k)$ at the current best estimate $\boldsymbol{\lambda}_k$ yields
\begin{eqnarray*}
\ell_{\rm M}(\boldsymbol{\lambda};\boldsymbol{y}) \equiv Q(\boldsymbol{\lambda}; \boldsymbol{\lambda}_k) - K(\boldsymbol{\lambda}; \boldsymbol{\lambda}_k),
\end{eqnarray*}
where
\begin{eqnarray}
Q(\boldsymbol{\lambda}; \boldsymbol{\lambda}_k) &=& E_{\pi(\boldsymbol{\beta} \mid \boldsymbol{Y}; \boldsymbol{\lambda}_k)}\left\{ \ell_{\rm P}(\boldsymbol{\beta}; \boldsymbol{Y}, \boldsymbol{\lambda}) +\cfrac{1}{2} \log\left| \boldsymbol{S}_{\boldsymbol{\lambda}} \right|_+ \right\}, \label{eq:Qdef} \\
K(\boldsymbol{\lambda}; \boldsymbol{\lambda}_k) &=& E_{\pi(\boldsymbol{\beta} \mid \boldsymbol{Y}; \boldsymbol{\lambda}_k)}\left\{\ell(\boldsymbol{\beta} \mid \boldsymbol{Y}; \boldsymbol{\lambda})\right\} = E_{\pi(\boldsymbol{\beta} \mid \boldsymbol{Y}; \boldsymbol{\lambda}_k)}\left\{\log \pi(\boldsymbol{\beta} \mid \boldsymbol{Y}; \boldsymbol{\lambda})\right\}. \nonumber
\end{eqnarray}
The E-step corresponds to the analytic calculation of the function $Q$, which is maximized with respect to $\boldsymbol{\lambda}$ at the M-step to provide $\boldsymbol{\lambda}_{k+1}$, as input for the next EM iteration. Using Jensen's inequality, direct calculation shows that $K(\boldsymbol{\lambda}; \boldsymbol{\lambda}_k) \leqslant K(\boldsymbol{\lambda}_k; \boldsymbol{\lambda}_k)$ for all $\boldsymbol{\lambda}$, and since $Q(\boldsymbol{\lambda}_{k+1};\boldsymbol{\lambda}_k) \geqslant Q(\boldsymbol{\lambda}_k; \boldsymbol{\lambda}_k)$, we have $\ell_{\rm M}(\boldsymbol{\lambda}_{k+1}; \boldsymbol{y}) \geqslant \ell_{\rm M}(\boldsymbol{\lambda}_k; \boldsymbol{y})$. Thus the EM algorithm transfers optimization of the log-marginal likelihood to that of $Q$, and ensures that $\ell_{\rm M}$ increases after every M-step. Under mild conditions, the algorithm is guaranteed to reach at least a local maximum \citep{EM77}. We first construct the function $Q$ used at the E-step.

\subsection{E-step}\label{sec:Estep}
\noindent Applying Bayes' rule to the posterior for $\boldsymbol{\beta}$, the non-trivial element of the function $Q$ in \eqref{eq:Qdef} is 
\begin{eqnarray}
E_{\pi(\boldsymbol{\beta} \mid \boldsymbol{Y}; \boldsymbol{\lambda}_k)}\left\{ \ell_{\rm P}(\boldsymbol{\beta}; \boldsymbol{Y}, \boldsymbol{\lambda}) \right\} &=&
\displaystyle \int \ell_{\rm P}(\boldsymbol{\beta}; \boldsymbol{y},\boldsymbol{\lambda})\ \pi(\boldsymbol{\beta} \mid \boldsymbol{Y}=\boldsymbol{y}; \boldsymbol{\lambda}_k) \, \mathrm{d}\boldsymbol{\beta}\nonumber \\
&=& \cfrac{\displaystyle \int\ell_{\rm P}(\boldsymbol{\beta}; \boldsymbol{y},\boldsymbol{\lambda}) \exp \ell_{\rm P}(\boldsymbol{\beta}; \boldsymbol{y},\boldsymbol{\lambda}_k) \, \mathrm{d}\boldsymbol{\beta}}{\displaystyle \int \exp\ell_{\rm P}(\boldsymbol{\beta}; \boldsymbol{y},\boldsymbol{\lambda}_k)\, \mathrm{d}\boldsymbol{\beta}}.\label{eq:EQ}
\end{eqnarray}
Both integrals are intractable, and as $\ell_{\rm P}$ may not be positive, the numerator cannot be expressed as the integral of an exponential function, which makes direct Laplace approximation impracticable. \cite{TierneyKassKadane1989} overcome this by approximating similar ratios using the moment generating function, as~\eqref{eq:EQ} is the expectation of a scalar function, $\ell_{\rm P}$, of the regression weights, seen as random variables with probability density their posterior. For any $\boldsymbol{\beta}$, let 
\begin{eqnarray*}
\ell_t(\boldsymbol{\beta}; \boldsymbol{y},\boldsymbol{\lambda}, \boldsymbol{\lambda}_k)=t\ell_{\rm P}(\boldsymbol{\beta}; \boldsymbol{y},\boldsymbol{\lambda})+\ell_{\rm P}(\boldsymbol{\beta}; \boldsymbol{y},\boldsymbol{\lambda}_k), \quad t \in \mathbb{R}.
\end{eqnarray*}
The conditional moment generating function of $\ell_{\rm P}(\boldsymbol{\beta};\boldsymbol{Y}, \boldsymbol{\lambda})$ is thus
\begin{eqnarray}
{\rm{M}}(t) = E_{\pi(\boldsymbol{\beta} \mid \boldsymbol{Y}; \boldsymbol{\lambda}_k)}\left[ \exp \left\{t \ell_{\rm P}(\boldsymbol{\beta};\boldsymbol{Y},\boldsymbol{\lambda}) \right\}\right] &=& \cfrac{\displaystyle \int \exp \ell_t(\boldsymbol{\beta};\boldsymbol{y},\boldsymbol{\lambda}, \boldsymbol{\lambda}_k) \, \mathrm{d}\boldsymbol{\beta}}{\displaystyle \int \exp \ell_0(\boldsymbol{\beta};\boldsymbol{y},\boldsymbol{\lambda}, \boldsymbol{\lambda}_k) \, \mathrm{d}\boldsymbol{\beta}}. \label{eq:MGF}
\end{eqnarray}
Expression \eqref{eq:MGF} is a ratio of two intractable integrals, each of which can be approximated using Laplace's method. Let 
\begin{eqnarray*}
\boldsymbol{\hat \beta}_t = \arg \max_{\boldsymbol{\beta}} \ell_t(\boldsymbol{\beta};\boldsymbol{y},\boldsymbol{\lambda}, \boldsymbol{\lambda}_k), \quad \boldsymbol{\hat \beta}_{k} = \arg \max_{\boldsymbol{\beta}} \ell_{\rm P}(\boldsymbol{\beta};\boldsymbol{y}, \boldsymbol{\lambda}_k) =\arg \max_{\boldsymbol{\beta}} \ell_0(\boldsymbol{\beta}; \boldsymbol{y},\boldsymbol{\lambda}, \boldsymbol{\lambda}_k)
\end{eqnarray*}
denote the maximizers of $\ell_t(\boldsymbol{\beta};\boldsymbol{y},\boldsymbol{\lambda}, \boldsymbol{\lambda}_k)$ and $\ell_{\rm P}(\boldsymbol{\beta};\boldsymbol{y},\boldsymbol{\lambda}_k)$, and write the negative Hessian matrix as
$H_t(\boldsymbol{\beta}; \boldsymbol{\lambda}, \boldsymbol{\lambda}_k) = tH_{\rm P}(\boldsymbol{\beta}; \boldsymbol{\lambda}) + H_{\rm P}(\boldsymbol{\beta}; \boldsymbol{\lambda}_k)$, where $H_{\rm P}$ is given in~\eqref{eq:Hp}. Second-order Taylor expansion of $\ell_t(\boldsymbol{\beta};\boldsymbol{y}, \boldsymbol{\lambda}, \boldsymbol{\lambda}_k)$ around $\boldsymbol{\hat \beta}_t$ yields the following approximation for the numerator of~\eqref{eq:MGF}
\begin{eqnarray*}
\displaystyle \int \exp{\ell_t(\boldsymbol{\beta};\boldsymbol{y}, \boldsymbol{\lambda}, \boldsymbol{\lambda}_k)}\,  \mathrm{d}\boldsymbol{\beta} & \approx & \exp{\ell_t(\boldsymbol{\hat \beta}_t;\boldsymbol{y}, \boldsymbol{\lambda}, \boldsymbol{\lambda}_k) } \ \displaystyle \int \exp{\left\{ -\cfrac{1}{2} (\boldsymbol{\beta}-\boldsymbol{\hat \beta}_t)^TH_t(\boldsymbol{\hat \beta}_t;\boldsymbol{\lambda}, \boldsymbol{\lambda}_k)(\boldsymbol{\beta}-\boldsymbol{\hat \beta}_t) \right\}}  \, \mathrm{d}\boldsymbol{\beta} \\
&=& \left(2\pi\right)^{p/2} \det{ {H_t(\boldsymbol{\hat \beta}_t;\boldsymbol{\lambda}, \boldsymbol{\lambda}_k)}}^{-1/2} \exp{\ell_t(\boldsymbol{\hat \beta}_t;\boldsymbol{y}, \boldsymbol{\lambda}, \boldsymbol{\lambda}_k) } + O(n^{-1}),
\end{eqnarray*}
where the determinant is well-defined because $H_t(\boldsymbol{\hat \beta}_t;\boldsymbol{\lambda}, \boldsymbol{\lambda}_k)$ is positive definite at convergence. On similarly applying Laplace approximation to the denominator of~\eqref{eq:MGF}, the conditional moment generating function becomes
\begin{eqnarray}
{\rm{M}}(t) &= & \exp{\left\{\ell_t(\boldsymbol{\hat \beta}_t;\boldsymbol{y},\boldsymbol{\lambda}, \boldsymbol{\lambda}_k)-\ell_{\rm P}(\boldsymbol{\hat \beta}_k;\boldsymbol{y}, \boldsymbol{\lambda}_k) \right\}} \cfrac{\det {H_t(\boldsymbol{\hat \beta}_t;\boldsymbol{\lambda}, \boldsymbol{\lambda}_k)}^{-1/2}}{\det {H_{\rm P}(\boldsymbol{\hat \beta}_k;\boldsymbol{\lambda}_k)}^{-1/2}} +O(n^{-2}),\label{eq:finMGF}
\end{eqnarray}
where the error is $O(n^{-2})$ rather than $O(n^{-1})$ because the error terms in the numerator and denominator almost cancel \citep[Theorem 1]{TierneyKassKadane1989}. The conditional expectation \eqref{eq:EQ} is obtained by differentiating \eqref{eq:finMGF} with respect to $t$ and evaluating it at $t=0$.

Whereas \cite{TierneyKassKadane1989} suggest numerical computation of such derivatives, we shall calculate them analytically. We need ${\rm{d}} \ell_t(\boldsymbol{\hat{\beta}}_t;\boldsymbol{\lambda}, \boldsymbol{\lambda}_k)/ {\rm{d}}t$ and ${\rm{d}}\det H_t(\boldsymbol{\hat{\beta}}_t;\boldsymbol{\lambda}, \boldsymbol{\lambda}_k)/{\rm{d}}t$, both evaluated at $t=0$. To simplify the notation we write ${\rm{d}}_0 \cdot / {\rm{d}}t$ to denote ${\rm{d}} \cdot / {\rm{d}}t \ |_{t=0}$ and similarly for $\partial_0 \cdot / \partial t$. 

\paragraph{Calculation of ${\rm{d}}_0 \ell_t(\boldsymbol{\hat{\beta}}_t;\boldsymbol{\lambda}, \boldsymbol{\lambda}_k)/ {\rm{d}}t$.}

As $\boldsymbol{\hat{\beta}}_t$ depends on $t$, 
\begin{eqnarray*}
\dfrac{\text{d}\ell_t}{\text{d}t}(\boldsymbol{\hat \beta}_t;\boldsymbol{y},\boldsymbol{\lambda}, \boldsymbol{\lambda}_k) = \ell_{\rm P}(\boldsymbol{\hat{\beta}}_t;\boldsymbol{y}, \boldsymbol{\lambda})+ t \cfrac{\partial \boldsymbol{\hat{\beta}}_t}{\partial t} \cdot  {\cfrac{\partial \ell_{\rm P}}{\partial \boldsymbol{\beta}}(\boldsymbol{\hat \beta}_t;\boldsymbol{y}, \boldsymbol{\lambda})} +  \cfrac{\partial \boldsymbol{\hat{\beta}}_t}{\partial t} \cdot {\cfrac{\partial \ell_{\rm P}}{\partial \boldsymbol{\beta}}(\boldsymbol{\hat \beta}_t; \boldsymbol{y},\boldsymbol{\lambda}_k)},
\end{eqnarray*}
where $\cdot$ denotes the scalar product. Since $\boldsymbol{\hat{\beta}}_t=\boldsymbol{\hat \beta}_k$ at $t=0$, and $\boldsymbol{\hat \beta}_k$ maximizes $\ell_{\rm P}(\boldsymbol{\beta}; \boldsymbol{y},\boldsymbol{\lambda}_k)$, we obtain
\begin{eqnarray}
{\dfrac{{\rm{d}}_{0} \ell_t}{{\rm{d}}t}(\boldsymbol{\hat{\beta}}_t;\boldsymbol{y},\boldsymbol{\lambda}, \boldsymbol{\lambda}_k)} &=&\ell_{\rm P}(\boldsymbol{\hat \beta}_k; \boldsymbol{y},\boldsymbol{\lambda}). \label{eq:dltilde}
\end{eqnarray}

\paragraph{Calculation of ${\rm{d}}_0\det{ H_t(\boldsymbol{\hat{\beta}}_t;\boldsymbol{\lambda}, \boldsymbol{\lambda}_k)/{\rm{d}}t }$.}

This requires $\partial_0 \boldsymbol{\hat{\beta}}_t / \partial t$, which we obtain by implicit differentiation of $\ell_t(\boldsymbol{\beta};\boldsymbol{y},\boldsymbol{\lambda}, \boldsymbol{\lambda}_k)$. At $\boldsymbol{\beta}=\boldsymbol{\hat{\beta}}_t$, we have ${\partial \ell_t(\boldsymbol{\beta};\boldsymbol{y}, \boldsymbol{\lambda},\boldsymbol{\lambda}_k)/\partial \boldsymbol{\beta}}\ |_{\boldsymbol{\beta}=\boldsymbol{\hat{\beta}}_t}=0$, so differentiating with respect to $t$ and setting $t=0$ yields
\begin{eqnarray}
U_{\rm P}(\boldsymbol{\hat{\beta}}_k; \boldsymbol{\lambda})-H_{\rm P}(\boldsymbol{\hat{\beta}}_k; \boldsymbol{\lambda}_k)\cdot {\cfrac{\partial_0 \boldsymbol{\hat{\beta}}_t}{\partial t}} =0.\label{eq:form}
\end{eqnarray}
As 
 $U_{\rm P}(\boldsymbol{\hat{\beta}}_k;\boldsymbol{\lambda})=U_{\rm P}(\boldsymbol{\hat{\beta}}_k; \boldsymbol{\lambda}_k)+ \boldsymbol{S}_{\boldsymbol{\lambda}_k} \boldsymbol{\hat{\beta}}_k -\boldsymbol{S}_{\boldsymbol{\lambda}}\boldsymbol{\hat{\beta}}_k=\left(\boldsymbol{S}_{\boldsymbol{\lambda}_k}-\boldsymbol{S}_{\boldsymbol{\lambda}} \right)\boldsymbol{\hat{\beta}}_k$, we get from \eqref{eq:form} that 
\begin{eqnarray}
{\cfrac{\partial_0 \boldsymbol{\hat{\beta}}_t}{\partial t}} &=& H^{-1}_{\rm P}(\boldsymbol{\hat{\beta}}_k; \boldsymbol{\lambda}_k)\left(\boldsymbol{S}_{\boldsymbol{\lambda}_k}-\boldsymbol{S}_{\boldsymbol{\lambda}} \right)\boldsymbol{\hat{\beta}}_k.\label{eq:betatilde}
\end{eqnarray}
Applying Jacobi's formula to ${\rm{d}}\det{H_t(\boldsymbol{\hat{\beta}}_t;\boldsymbol{\lambda}, \boldsymbol{\lambda}_k)}/{\rm{d}}t$ and evaluating the result at $t=0$ yields 
\begin{eqnarray}
\dfrac{{\rm{d}}_0\det{H_t(\boldsymbol{\hat{\beta}}_t;\boldsymbol{\lambda}, \boldsymbol{\lambda}_k)} }{{\rm{d}}t} = \det{H_{\rm P}(\boldsymbol{\hat{\beta}}_k;\boldsymbol{\lambda}_k)} \times \Tr \left[H_{\rm P}^{-1}(\boldsymbol{\hat{\beta}}_k;\boldsymbol{\lambda}_k)  \left\{H_{\rm P}(\boldsymbol{\hat{\beta}}_k; \boldsymbol{\lambda}) + { \cfrac{{\rm{d}}_0 H}{{\rm{d}}t}(\boldsymbol{\hat{\beta}}_t)}\right\} \right],\label{eq:dHt}
\end{eqnarray}
where the last derivative term can be computed by the chain rule and using~\eqref{eq:betatilde}. On inserting~\eqref{eq:dHt} and \eqref{eq:dltilde} into the derivative of~\eqref{eq:finMGF} with respect to $t$ and evaluating the result at $t=0$, we find after a little algebra that 
\begin{eqnarray*}
Q(\boldsymbol{\lambda}; \boldsymbol{\lambda}_k) &\equiv &   - \cfrac{1}{2} \ \Tr \left[H^{-1}_{\rm P}(\boldsymbol{\hat{\beta}}_k; \boldsymbol{\lambda}_k) \left\{H_{\rm P}(\boldsymbol{\hat{\beta}}_k; \boldsymbol{\lambda}) + {\cfrac{{\rm{d}}_0 H}{{\rm{d}}t} (\boldsymbol{\hat{\beta}}_t)} \right\} \right]\nonumber \\
& & \qquad +\cfrac{1}{2} \log \left| \boldsymbol{S}_{\boldsymbol{\lambda}} \right|_+   +\ell_{\rm P}(\boldsymbol{\hat{\beta}}_k; \boldsymbol{y}, \boldsymbol{\lambda}) + O(n^{-2}).
\end{eqnarray*}
The order $O(n^{-2})$ of the error in $Q$ over the usual $O(n^{-1})$ error for Laplace approximation shows that this E-step provides a potentially better approximation to the function to be maximized to obtain the smoothing parameters. Moreover, the proposed approach is clearly an outer iteration optimization, since $Q$ is defined in terms of the maximum $\boldsymbol{\hat \beta}_k$ rather than its intermediate estimate, as in the performance iteration optimization; see Section~\ref{sec:litrev}. This guarantees that the smoothing parameters will converge to a local maximizer of the log-marginal likelihood. As we shall now see, this approximate E-step greatly simplifies the M-step; the crux is that $\boldsymbol{\hat \beta}_k$ depends by definition on $\boldsymbol{\lambda}_k$ alone, and not on $\boldsymbol{\lambda}$. 

\subsection{M-step} \label{subsec:Mstep}
The M-step entails the calculation of the gradient and Hessian matrix of $Q$ with respect to the smoothing parameters $\boldsymbol{\lambda}$. We first show that the derivative of $\partial_0 \boldsymbol{\hat{\beta}}_t/\partial t$ in \eqref{eq:betatilde} with respect to $\boldsymbol{\lambda}$ equals $\partial\boldsymbol{\hat{\beta}}_k/\partial \boldsymbol{\lambda}_k$. As $\boldsymbol{\hat{\beta}}_k$ is the solution to the equation $U_{\rm P}(\boldsymbol{\hat{\beta}}_k; \boldsymbol{\lambda}_k)=0$, taking the derivative with respect to the $j$-th component $\lambda_{k,j}$ of $\boldsymbol{\lambda}_k$ yields
\begin{eqnarray}
\cfrac{\partial \boldsymbol{\hat{\beta}}_k}{\partial \lambda_{k,j}} = -H^{-1}_{\rm P}(\boldsymbol{\hat{\beta}}_k; \boldsymbol{\lambda}_k) S_j \boldsymbol{\hat{\beta}}_k =\cfrac{\partial \partial_0 \boldsymbol{\hat{\beta}}_t}{\partial \lambda_j \partial t},\label{eq:c0}
\end{eqnarray}
since $\partial {\boldsymbol S}_{\boldsymbol{\lambda}_{k,j}}/\partial\lambda_{k,j} =S_j=\partial {\boldsymbol S}_{\boldsymbol{\lambda}}/\partial\lambda_j $. Using the chain rule, equality~\eqref{eq:c0} implies that 
\begin{eqnarray}
\cfrac{\partial H}{\partial \lambda_{k,j}}  (\boldsymbol{\hat{\beta}}_k) = \cfrac{\partial {\rm{d}}_0 H}{\partial \lambda_j {\rm{d}}t} (\boldsymbol{\hat{\beta}}_t).\label{eq:ddH}
\end{eqnarray} 
Let $\boldsymbol{\hat{\beta}}^{(j)}_k$ denote the block of $\boldsymbol{\hat{\beta}}_k$ corresponding to $S_j$ and the smooth function $f_j$. Using~\eqref{eq:ddH}, the components of the gradient of the E-step are
\begin{eqnarray}
G_j(\boldsymbol{\lambda}; \boldsymbol{\lambda}_k) &=& \cfrac{\partial Q}{\partial \lambda_j} (\boldsymbol{\lambda}; \boldsymbol{\lambda}_k) = \cfrac{1}{2} \left\{\Tr \left( \boldsymbol{S}_{\boldsymbol{\lambda}}^{-} S_j\right) -  c_{k,j} \right\}, \label{eq:G}
\end{eqnarray}
where 
\begin{eqnarray*}
c_{k,j} = {\boldsymbol{\hat{\beta}}^{(j)}_k}^{T} S_j \boldsymbol{\hat{\beta}}^{(j)}_k +\Tr \left[H^{-1}_{\rm P}(\boldsymbol{\hat{\beta}}_k; \boldsymbol{\lambda}_k)\left\{ S_j + \cfrac{\partial H}{\partial \lambda_{k,j}}  (\boldsymbol{\hat{\beta}}_k) \right\} \right] \in \mathbb{R}. 
\end{eqnarray*}
By construction in~\eqref{eq:SblockDiag}, $\boldsymbol{S}_{\boldsymbol{\lambda}}$ is a block-diagonal matrix whose blocks are of the general form $S=\lambda_jS_j$, which implies that
$\Tr ( \boldsymbol{S}_{\boldsymbol{\lambda}}^{-} S_j)={\rm{rank}}(S_j)/\lambda_j$ and yields the closed form
\begin{eqnarray}
\hat{\lambda}_{k+1, j}= \cfrac{{\rm{rank}}(S_j)}{c_{k,j}}, \quad j=1, \ldots, q, \label{eq:lamb}
\end{eqnarray}
where $c_{k,j}>0$ is always true by positivity of the smoothing parameters. The Hessian matrix of $Q$ is therefore diagonal with negative elements $-{\rm{rank}}(S_j)/(2\lambda_j^2)<0$, so \eqref{eq:lamb} are always maximizers. The corresponding components of $\boldsymbol{\lambda}$ are positive, so it might be thought necessary to set $\boldsymbol{\rho}=\log \boldsymbol{\lambda}$ componentwise before the approximate EM optimization and then back-transform afterwards. This would have led to finding the roots of 
\begin{eqnarray*}
\tilde{G}_j(\boldsymbol{\rho}; \boldsymbol{\rho}_k)= \cfrac{\partial Q}{\partial \rho_j} (\exp \boldsymbol{\rho}; \exp \boldsymbol{\rho}_k) = G_j(\boldsymbol{\lambda}; \boldsymbol{\lambda}_k) \exp \rho_j,
\end{eqnarray*}
which are also the roots of $G_j$ in~\eqref{eq:G}, with Hessian components $-(c_{k,j}\exp \rho_j)/2<0$, so the positivity constraint need not be explicitly included. The diagonality of the Hessian matrix of $Q$ allows embarrassingly parallel computation of the M-step, which provides substantial speed when $q$, the number of smooth functions, is large.

Overall, the $k$-th iteration of the approximate EM algorithm consists in
\begin{enumerate}
\item[1)] using the current best estimate $\boldsymbol{\lambda}_k$ to maximize the penalized log-likelihood~\eqref{eq:lp} to get $\boldsymbol{\hat \beta}_k$;
\item[2)] computing $\boldsymbol{\lambda}_{k+1}$, possibly in parallel, using \eqref{eq:lamb};
\item[3)] updating $k+1$ to $k$.
\end{enumerate}
Learning of the regression weights is incorporated into step 1), which is based on a Newton--Raphson algorithm. Given the trial value $\boldsymbol{\beta}_{l}$, each iteration involves 
\begin{itemize}
\item[a)] making $H_{\rm P}(\boldsymbol{\beta}_l;\boldsymbol{\lambda}_k)$ positive definite;
\item[b)] evaluating the updating step
\begin{eqnarray*}
\boldsymbol{\beta}_{l+1} = \boldsymbol{\beta}_{l} + \gamma \Delta_k, \quad \Delta_l=H_{\rm P}^{-1}(\boldsymbol{\beta}_l;\boldsymbol{\lambda}_k)U_{\rm P}(\boldsymbol{\beta}_l;\boldsymbol{\lambda}_k),
\end{eqnarray*}
\end{itemize}
where $\gamma$ is the learning rate. At step a), the positive definiteness of $H_{\rm P}(\boldsymbol{\beta}_l;\boldsymbol{\lambda}_k)$ is guaranteed by increasing eigenvalues smaller than a certain positive tolerance to that tolerance. The stability of the algorithm is ensured by successively halving $\gamma$ at step b) until the penalized log-likelihood increases. At convergence, $\boldsymbol{\hat{\beta}}_k=\boldsymbol{\beta}_{l+1}$, and the identifiability of the regression weights must be checked to ensure that $H_{\rm P}(\boldsymbol{\hat \beta}_k; \boldsymbol{\lambda}_k)$ is invertible, since this matrix is required for calculating the smoothing parameters. By definition, the regression model is identifiable if and only if its weights are linearly independent, so a strategy for dealing with lack of identifiability is to keep only the ${\rm{rank}}\ H_{\rm P}(\boldsymbol{\hat \beta}_k; \boldsymbol{\lambda}_k)=r \leqslant p$ linearly independent regression weights. An efficient and stable method to reveal these is QR decomposition with column pivoting \citep[\S~5.4.2]{golub}. The QR factorization finds a permutation matrix $P \in \mathbb{R}^{p\times p}$ such that $H_{\rm P}(\boldsymbol{\hat \beta}_k; \boldsymbol{\lambda}_k)P=QR$, where the first $r$ columns of $Q$ form an orthonormal basis for $H_{\rm P}$. As the permutation matrix tracks the moves of the columns of $H_{\rm P}$, the $r$ identifiable weights are the first $r$ components of the re-ordered vector $P^T\hat{\boldsymbol{\beta}}$. The remaining $p-r$ weights are hence linearly dependent, and should be excluded from the model, together with the corresponding columns of $\boldsymbol{X}$, and the rows and columns of $\boldsymbol{S}_{\boldsymbol{\lambda}_k}$.

Steps 1)--3) are iterated until the gradient of the log-marginal likelihood is sufficiently small. \citet{Oakes1999} showed that this gradient can be written in terms of that of $Q$, as $\partial \ell_{\rm M}(\boldsymbol{\lambda};\boldsymbol{y})/\partial \boldsymbol{\lambda} = G(\boldsymbol{\lambda};\boldsymbol{\lambda}_k) |_{\boldsymbol{\lambda}_k=\boldsymbol{\lambda}}$. Since $G(\boldsymbol{\lambda}_{k+1}; \boldsymbol{\lambda}_k)=0$, the convergence criterion is equivalent to checking that for each $j$, 
\[
G_j(\lambda_{k+1,j})=\cfrac{1}{2} (c_{k,j}-c_{k+1,j}) < \epsilon,
\]
where $\epsilon$ is a small tolerance. Furthermore, the diagonality of the Hessian of $Q$ allows one to check convergence independently for each smoothing parameter, so that only unconverged ones must be updated. In practice, the smoothing parameters may be large enough that significant changes in some components of $\boldsymbol{\lambda}$ yield insignificant changes of the penalized log-likelihood, which suggests deeming convergence when there is no significant change in the penalized log-likelihood. The full optimization is summarized in the three-step iteration, whose leading computational costs in the worst-case scenario are $O(np^2)$ for the computation of the Hessian of the log-likelihood, $O(p^3)$ for its inversion, and $O(np^2)$ for its derivative.

The EM algorithm provides an elegant and straightforward approach to maximization of the log-marginal likelihood. We obtained an accurate E-step based on the approximation of~\cite{TierneyKassKadane1989} with error $O(n^{-2})$, and derived a closed form for the M-step that circumvents evaluation of the expensive and numerically unstable function $Q$. This indirect approach leads to an important simplification of the learning procedure compared to the direct Laplace approach. As the M-step is always upward, no learning rate tuning is required: there is no need for intermediate evaluation of the log-marginal likelihood or its Hessian matrix. The former circumvents inner optimizations of the penalized log-likelihood and evaluation of unstable terms when the components of $\boldsymbol{\lambda}$ differ in magnitude, and the latter avoids computation of the fourth-order log-likelihood derivatives, which may be difficult to calculate, computationally expensive and numerically unstable~\citep{Wood2011,Wood2016}. Moreover, the diagonality of the Hessian matrix of $Q$ allows parallelization of the M-step and update of the unconverged smoothing parameters only, providing thus an additional shortcut.  We assess the performance of the proposed methodology in Section~\ref{sec:simulation}.


\section{Simulation study} \label{sec:simulation}
\begin{figure}[!t]
  \centering
  \begin{minipage}[b]{0.4\textwidth}
    \includegraphics[scale=.4]{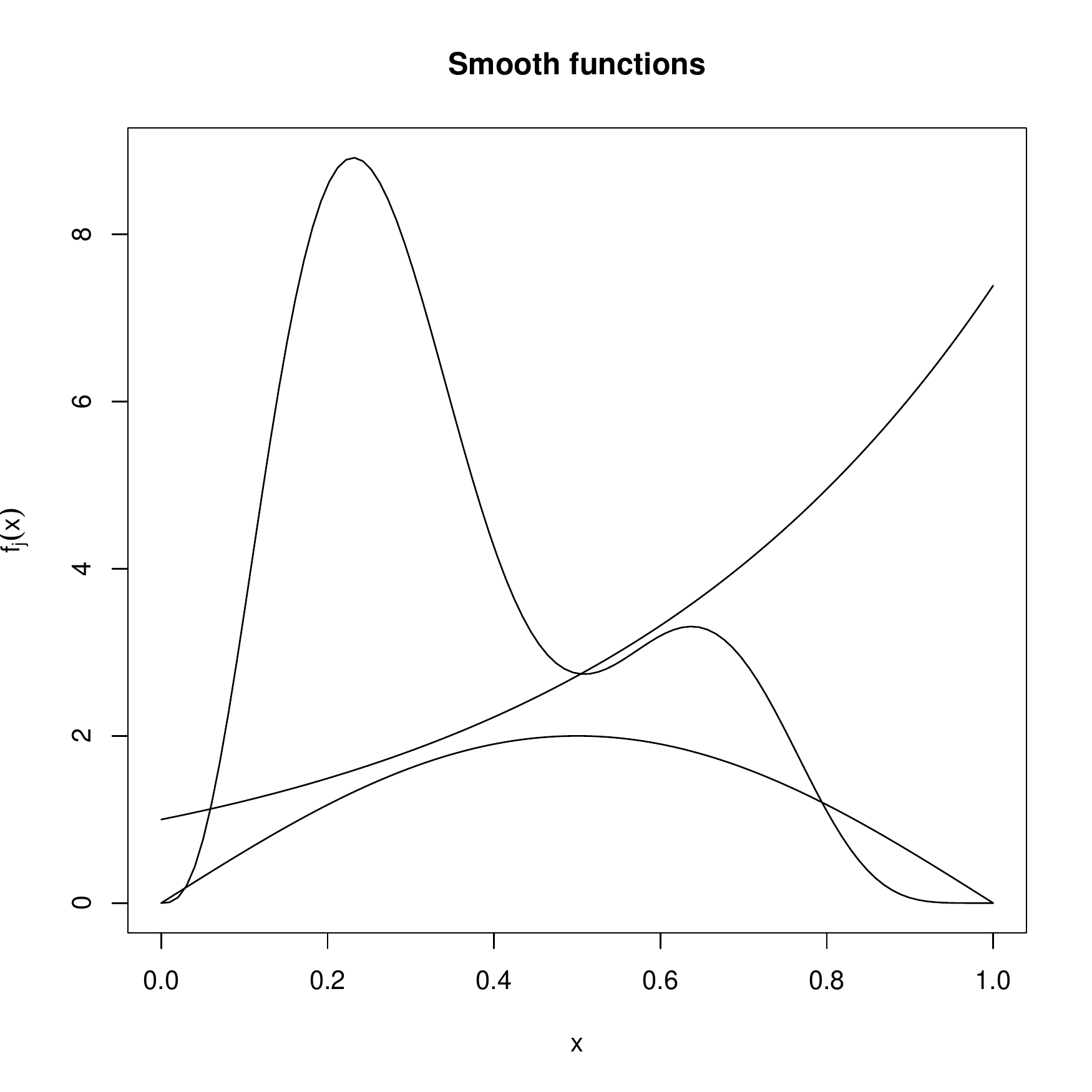}
    \caption{Original $f_j$ for $j=1, \ldots, 3$.}
    \label{fig:sim1}
  \end{minipage}
  \hfill
  \begin{minipage}[b]{0.4\textwidth}
    \includegraphics[scale=.4]{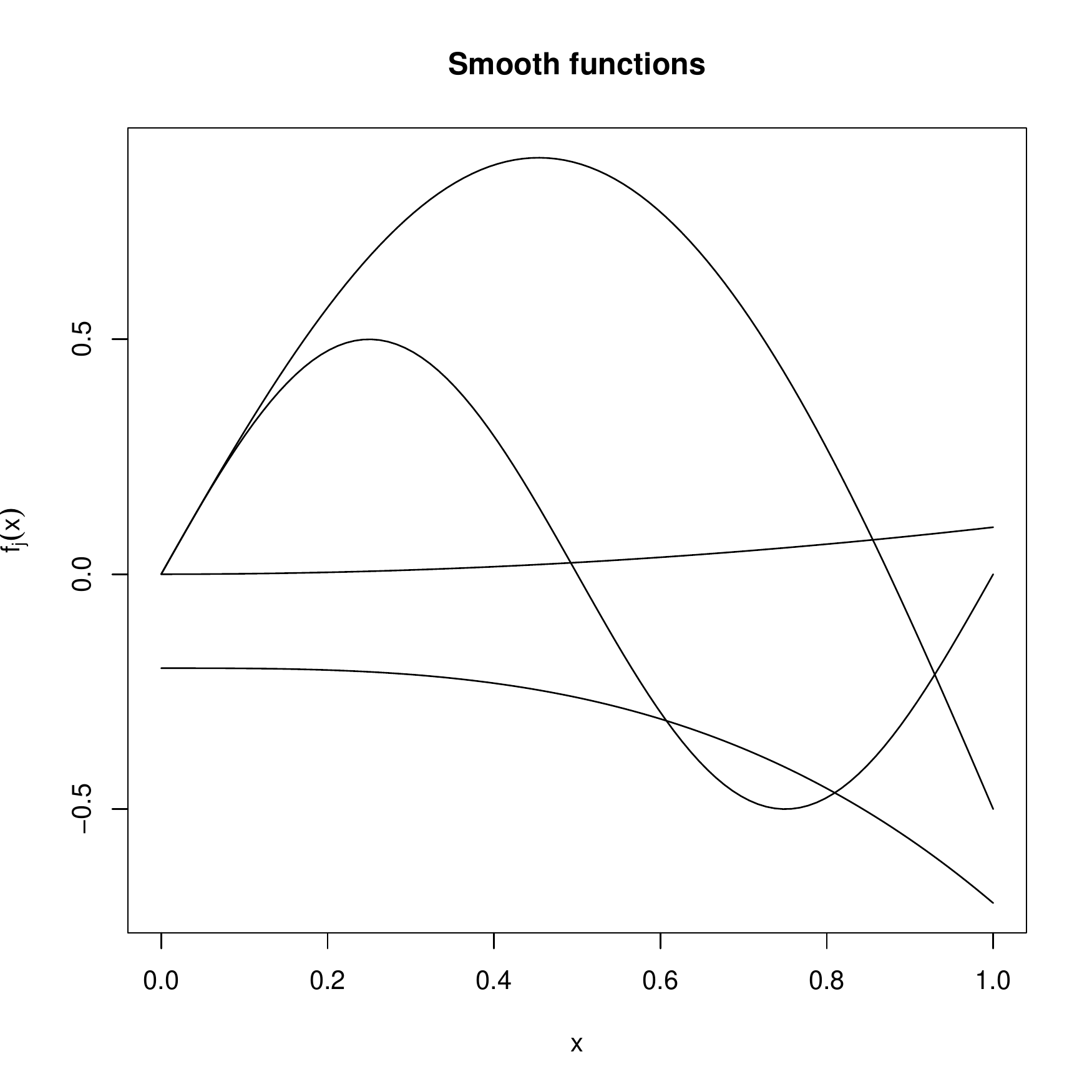}
    \caption{Original $f_j$ for $j=4, \ldots, 7$.}
    \label{fig:sim2}
  \end{minipage}
\end{figure}

We generated $R=100$ replicates of training sets of $n=25000$ examples from a variety of probability distributions with parameters that depend on smooth functions of inputs. Let $x_1, \ldots, x_7$ be independent vectors of $n$ identically distributed standard uniform variables. Figures~\ref{fig:sim1} and~\ref{fig:sim2} illustrate the seven smooth functions we considered
\begin{eqnarray*}
f_1(x) &=& 10^4 x^3(1-x)^6 \left\{(1-x)^{4} + 20 x^8 \right\}, \quad
f_2(x) = 2\sin(\pi x), \quad f_3(x) = \exp(2x), \\
f_4(x) &=& 0.1x^2, \quad f_5(x) = \sin(2\pi x)/2, \quad f_6(x) = -0.2-x^3/2, \quad f_7(x) = -x^2/2 + \sin(\pi x).
\end{eqnarray*}

\noindent With the functional parameters
\begin{eqnarray*}
\mu(x_1, x_2, x_3) &=& \sum_{j=1}^3f_j(x_j), \quad
\sigma(x_4, x_5, x_6) = \sum_{j=4}^6f_j(x_j),\quad \xi(x_7) = f_7(x_7),
\end{eqnarray*}
we generated $n$ training examples from the following distributions:
\begin{itemize}
\item Gaussian distribution with mean $\mu(x_1, x_2, x_3)$ and standard deviation $\exp\{\sigma(x_4, x_5, x_6)/2\}$,
\item Poisson distribution with rate $\exp\{\mu(x_1, x_2, x_3)/6\}$,
\item Exponential distribution with rate $\exp\{\mu(x_1, x_2, x_3)/6\}$,
\item Gamma distribution with shape $\exp\{\mu(x_1, x_2, x_3)/6\}$ and scale $\exp\{-\sigma(x_4, x_5, x_6)\}$, 
\item Binomial distribution with probability of success $1/[1+\exp \{-\mu(x_1, x_2, x_3)+5\}/6]$,
\item Generalized extreme value (GEV) distribution with location $\mu(x_1, x_2, x_3)$, scale $\exp \sigma(x_4, x_5, x_6)$ and shape $\xi(x_7)$; see Section~\ref{subsec:GEV} for further details.
\end{itemize}

\begingroup
\setlength{\tabcolsep}{10pt} 
\renewcommand{\arraystretch}{1.2}
\begin{table}[!t]
\caption{Means ($\times 10^{-2}$) over 100 replicates of the integrated mean squared errors of the learned functional parameters for a variety of models and \texttt{R} packages. The variances ($\times 10^{-6}$) appear as subscripts. \label{tab:MSE1}}
\centering
\begin{tabular}{ l l rrr}
\hline
Model & Package & $\boldsymbol{\hat{\mu}}$ & $\boldsymbol{\hat{\sigma}}$ & $\boldsymbol{\hat{\xi}}$  \\ 
\hline
Gauss & \verb|multgam| & $2.47_{2.93}$ & $0.04_{0.01}$ & $-$ \\ 
& \verb|mgcv| gam & $2.47_{2.93}$ & $0.04_{0.01}$ & $-$ \\ 
\hline
Poisson & \verb|multgam| & $1.63_{4.56}$ & $-$ & $-$\\ 
& \verb|mgcv| gam  & $1.61_{4.67}$ & $-$ & $-$ \\ 
& \verb|mgcv| bam  & $1.62_{7.33}$ & $-$ & $-$ \\ 
& \verb|INLA| & $9.91_{7.48}$ & $-$ & $-$ \\ 
\hline
Exponential & \verb|multgam| & $3.67_{114.02}$ & $-$ & $-$ \\ 
& \verb|mgcv| gam & $3.75_{115.26}$ & $-$ & $-$ \\ 
& \verb|mgcv| bam & $3.69_{123.74}$ & $-$ & $-$  \\ 
& \verb|INLA| & $11.95_{87.06}$ & $-$ & $-$ \\ 
\hline
Gamma & \verb|multgam|  & $1.78_{9.27}$ & $0.02_{0.01}$ & $-$ \\ 
\hline
Binomial & \verb|multgam|  & $38.51_{0.99}$ & $-$ & $-$ \\ 
& \verb|mgcv| gam  & $38.51_{0.99}$ & $-$ & $-$ \\ 
& \verb|mgcv| bam  & $38.51_{0.99}$ & $-$ & $-$\\ 
& \verb|INLA| & $38.51_{0.99}$ & $-$ & $-$ \\ 
\hline
GEV & \verb|multgam| & $3.58_{8.61}$ & $0.15_{0.11}$ & $0.41_{1.06}$\\ 
& \verb|mgcv| gam & $3.63_{8.56}$ & $0.27_{77.63}$ & $0.67_{336.54}$\\ 
\hline
\end{tabular}
\end{table}
\endgroup

We fit the six models using cubic regression splines with evenly spaced knots in the predictor range values. We used ten basis functions for each of the smooth functions $f_j$. 
We computed the integrated mean squared error between the true and learned functional parameters, represented by hats, for each of the $r$ replicates
\begin{eqnarray*}
{\rm MSE}(\boldsymbol{\hat \theta}^{(d)[r]}) = \cfrac{1}{n}\sum_{i=1}^n  
\left(\theta_i^{(d)[r]} - \hat{\theta}_i^{(d)[r]}\right)^2,
\end{eqnarray*}
where $\boldsymbol{\hat \theta}^{(d)}$ is $\boldsymbol{\hat \mu}$, $\boldsymbol{\hat \sigma}$ or $\boldsymbol{\hat \xi}$. Table~\ref{tab:MSE1} summarizes the results for the proposed approach, \verb|multgam|, and three state-of-the-art methods implemented in the \verb|R| packages \verb|mgcv| gam~\citep{Wood2011, Wood2016}, \verb|mgcv| bam~\citep{Wood2015}, and \verb|INLA|~\citep{inla}. We also tried both Stan algorithms~\citep{stan}, fully Bayesian approach with Markov Chain Monte Carlo sampling and approximate variational Bayes, through the \verb|R| package \verb|brms| \citep{brms}, but a single replicate for a single functional parameter model run with four cores took five and three hours respectively, so the full simulation study would have taken much more than four months, which is infeasible. Another widely used \verb|R| package, \verb|VGAM|, does not offer automatic smoothing, and choosing $\boldsymbol{\lambda}$ manually for each $f_j$ for each model would have been tedious and error-prone. Use of the \verb|R| package \verb|gamlss| turned out to be infeasible. Some results for the Gauss, Gamma and GEV models are missing from Table~\ref{tab:MSE1} because the corresponding packages do not support them. Moreover, \verb|multgam| failed on 17 replicates for the GEV model, whereas \verb|mgcv| gam failed on 46 replicates, so the values shown are based on 83 and 54 training sets respectively. Table~\ref{tab:MSE1} shows that \verb|multgam| is the only package which supports all the classical models, and its small errors and low variances demonstrate the high accuracy and reliability of its estimates. The proposed method is competitive with both methods in \verb|mgcv|, whereas \verb|INLA| is less accurate. The new method is considerably better for the GEV model; it could fit $83$ of the replicates, compared to $54$ for \verb|mgcv| gam, and the estimates themselves were more accurate and less variable. The only model where all the methods give equally poor results is the binomial.

\begingroup
\setlength{\tabcolsep}{10pt} 
\renewcommand{\arraystretch}{1.2}
\begin{table}[!t]
\caption{Timing (s) for a variety of models and three training set sizes $n$. The notation $x^y$ means $x \times 10^y$. The notation $t^{(\times)}$ means the computation failed to converge after $t$ seconds, and $\times$ and $?$ indicate respectively failure to converge and that computations are still running at the time of submission. The ratios $R_{x^y}$ are with respect to \texttt{multgam}, which does not benefit from the parallelization of the M-step.
 \label{tab:timing}}
 \centering
 \begin{footnotesize}
\begin{tabular}{ l l rrr | rrr}
\hline
Model & Package & $2.5^4$ & $1^5$ & $5^5$ & $R_{2.5^4}$ & $R_{1^5}$ & $R_{5^5}$ \\ 
\hline
Gauss & \verb|multgam| & $3.38$ & $34.87$ & $93.85$ & $1$ & $1$ & $1$ \\
& \verb|mgcv| gam & $51.22$ & $549.88$ & $3861.33$ & $15.15$ & $15.77$ & $41.14$ \\
& \verb|brms| MCMC & $?$ & $359140^{(\times)}$ & $387242^{(\times)} $ & $?$ & $\times $ & $\times $ \\
& \verb|brms| VB & $?$ & $384702^{(\times)} $ & $394031^{(\times)} $ & $?$ & $\times $ & $\times $ \\
\hline
Poisson & \verb|multgam| & $0.33$ & $4.05$ & $16.39$ & $1$ & $1$ & $1$ \\
& \verb|mgcv| gam  & $4.25$ & $60.27$ & $2157.61$ & $12.88$ & $14.88$ & $131.64$ \\
& \verb|mgcv| bam  & $1.37$ & $9.83$ & $30.30$ & $4.15$ & $2.43$ & $1.85$ \\
& \verb|INLA| & $459.71$ & $12077.35$ & $\times $ & $1393.06$ & $2982.06$ & $\times $ \\ 
& \verb|brms| MCMC & $?$ & $?$ & $?$ & $?$ & $?$ & $?$ \\
& \verb|brms| VB & $?$ & $?$ & $?$ & $?$ & $?$ & $?$ \\
\hline
Exponential & \verb|multgam| & $0.71$ & $5.03$ & $19.89$ & $1$ & $1$ & $1$ \\
& \verb|mgcv| gam  & $4.67$ & $56.55$ & $340.15$ & $6.58$ & $11.24$ & $17.10$ \\
& \verb|mgcv| bam  & $1.49$ & $6.83$ & $33.13$ & $2.10$ & $1.36$ & $1.67$ \\
& \verb|INLA| & $466.77$ & $\times $ & $\times $ & $657.42$ & $\times $ & $\times $ \\ 
& \verb|brms| MCMC & $?$ & $?$ & $?$ & $?$ & $?$ & $?$ \\
& \verb|brms| VB & $?$ & $?$ & $?$ & $?$ & $?$ & $?$ \\ 
\hline
Gamma & \verb|multgam| & $2.67$ & $30.75$ & $97.04$ & $1$ & $1$ & $1$ \\
\hline
Binomial & \verb|multgam| & $0.38$ & $7.90$ & $19.27$ & $1$ & $1$ & $1$ \\
& \verb|mgcv| gam & $3.33$ & $58.51$ & $463.46$ & $8.76$ & $7.41$ & $24.05$ \\
& \verb|mgcv| bam & $1.23$ & $8.81$ & $22.60$ & $3.24$ & $1.12$ & $1.17$ \\
& \verb|INLA| & $299.91$ & $11543.32$ & $ \times$ & $789.24$ & $1461.18$ & $\times $ \\ 
& \verb|brms| MCMC & $?$ & $?$ & $?$ & $?$ & $?$ & $?$ \\
& \verb|brms| VB & $?$ & $?$ & $?$ & $?$ & $?$ & $?$ \\
\hline
GEV & \verb|multgam| & $8.24$ & $440.22$ & $\times $ & $1$ & $1$ & $1$ \\
& \verb|mgcv| gam & $167.13$ & $\times $ & $\times $ & $20.28$ & $\times $ & $\times $ \\
& \verb|brms| MCMC & $?$ & $?$ & $?$ & $?$ & $?$ & $?$ \\
& \verb|brms| VB & $?$ & $?$ & $?$ & $?$ & $?$ & $?$ \\
\hline
\end{tabular}
\end{footnotesize}
\end{table}
\endgroup

Table~\ref{tab:timing} gives a timing comparison for training sets of different sizes generated from the models described above. The computations were performed on a $2.80$ GHz Intel i7-7700HQ laptop using Ubuntu. The proposed method is always the fastest, more so for large training sets, and substantially outperforms \verb|mgcv| gam and \verb|INLA|. Moreover, it can fit the GEV model at sizes unmatched by existing software. The package \verb|INLA| fails with a half-million observations for all the models. Rather surprisingly, the proposed method is faster than \verb|multgam| bam, which is specifically designed for large datasets and exploits parallel computing, whereas \verb|multgam| performs the M-step serially for fair comparisons. Furthermore, the speed of \verb|mgcv| bam should be balanced by lack of reliability of its performance iteration algorithm; see Section~\ref{sec:litrev}. Table~\ref{tab:timing} demonstrates that speed and reliability need not be exclusive. One reason why \verb|mgcv| gam is slow is that it evaluates the fourth-order log-likelihood derivatives. Except for the GEV model, these are not difficult to compute, but they seem to entail significant overhead, evidenced by the difference in performance between \verb|multgam| and \verb|mgcv| gam. Overall, the new approach gives a substantial gain in speed with no loss in accuracy, and in some cases, it is the sole approach feasible. In Section~\ref{sec:dataAnalys} we apply the proposed method to environmental extreme data. 

\section{Data analysis}\label{sec:dataAnalys}
We analyze monthly maxima of temperature, which are non-stationary and using stationary models to make inference about them results in underestimation of risk, with serious potential consequences for human lives and insurance companies. The generalized extreme-value distribution, widely used for modeling maxima and minima, will serve as our underlying probability model.

\subsection{Model}\label{subsec:GEV}
Let $y_1, \ldots, y_n$ be the maxima of blocks of observations from an unknown probability distribution. Extreme value theory~\citep{FisherTippett1928, HaanFerreira} implies that as the block size increases and under mild conditions, each of the $y_i$ follows a GEV$(\mu_i, \sigma_i, \xi_i)$ distribution with parameters the location $\mu_i \in \mathbb{R}$, the scale $\sigma_i>0$ and the shape $\xi_i  \in \mathbb{R}$,
\begin{eqnarray*}
{F(y_i;\mu,\sigma_i,\xi_i)} &=& \begin{cases}
\exp\left[-\left\{1+\xi_i \left(\cfrac{y_i-\mu_i}{\sigma_i} \right) \right\}^{-1/\xi_i}_+ \right], & \xi_i \neq 0, \\
\exp\left[-\exp \left\{-\left(\cfrac{y_i-\mu_i}{\sigma_i} \right) \right\} \right], & \xi_i = 0,
\end{cases} 
\end{eqnarray*} 
where $a_+=\max(a,0)$. This encompasses the three classical models for maxima \citep{Jenkinson1955}: if $\xi_i>0$, the distribution is Fr\'echet; if $\xi_i<0$, it is reverse Weibull; and if $\xi_i\to0$, it is Gumbel. The shape parameter is particularly important since it controls the tail properties of the distributions. The expectation of $Y_i$ is
\begin{eqnarray}
E(Y_i)=
\begin{cases}
\mu_i +  \cfrac{\sigma_i}{\xi_i}\left\{\Gamma(1-\xi_i)-1\right\}, &  \xi_i \neq 0, \ \xi_i <1,\\
\mu_i + \gamma\sigma_i, &  \xi_i = 0,\\
\infty, &  \xi_i \geqslant 1,\\
\end{cases} \label{eq:MEAN}
\end{eqnarray}
where $\gamma$ is Euler's constant. Non-stationarity of~\eqref{eq:MEAN} could stem from changes in any of the parameters, and as intepretability is priority in risk assessment, a multiple GAM model for the GEV distribution is well justified.

Most data analyses involving non-stationary extremes use a parametric or semi-parametric form in the location and/or scale parameters while keeping the shape a fixed scalar \citep[\S4]{chavez2012}, even though it may be plausible that it varies---seasonal effects, for example, may stem from different physical processes with different extremal behaviors. Fixing the shape parameter is a pragmatic choice driven by the difficulty of learning it from limited data in a numerically stable manner. The only paper learning a functional shape parameter for extremes is \citet{Chavez2005} in the context of the generalized Pareto distribution, but their approach involves manual tuning of the smoothing parameters and has some drawbacks. First, training is based on backfitting, whose limitations were outlined in Section~\ref{sec:litrev}. Second, the optimization is in the spirit of performance iteration, with one updating step for the smoothing followed by another for the regression model; drawbacks of this were also discussed in Section~\ref{sec:litrev}. Third, optimization is sequential rather than simultaneous, by alternating a regression step for each smooth term when there are several and alternating backfitting steps for each functional parameter separately. Fourth, convergence may only be guaranteed when the functional parameters are orthogonal, meaning that the methodology may not extend to more than two. Moreover, the smoothing method is applied to orthogonalized distribution parameters that may be awkward to interpret. To illustrate our methodology, we learn a functional shape in a generic and stable manner; this is of separate interest for the modeling of non-stationary extremes. 

In our earlier general terms, $\theta_i^{(1)}=\mu_i$, $\theta_i^{(2)}=\exp \tau_i$, where $\tau_i=\log \sigma_i$ to ensure positivity of the scale, and $\theta_i^{(3)}=\xi_i$. Let $\Omega$ and $\Omega_0$ denote the partition of the support as 
\begin{eqnarray*}
 \Omega &=& \left\{y_i \in \mathbb{R}: \xi_i >0, y_i >\mu_i-\exp \tau_i /\xi_i \right\} \cup \left\{y_i \in \mathbb{R} : \xi_i <0, y_i <\mu_i-\exp \tau_i/\xi_i \right\}, \\
\Omega_0 &=& \left\{(y_i, \xi_i): \ y_i \in \mathbb{R}, \xi_i = 0 \right\}.
\end{eqnarray*}
The corresponding log-likelihood is then
\begin{eqnarray*}
\ell_{\rm L}(\boldsymbol{\mu}, \boldsymbol{\tau}, \boldsymbol{\xi}; \boldsymbol{y}) &=& \sum_{i=1}^n \ell_{\rm L}^{(i)}(\mu_i, \tau_i, \xi_i; y_i),
\end{eqnarray*} 
where the individual contributions are
\begin{eqnarray*}
 \ell_{\rm L}^{(i)}(\mu_i, \tau_i, \xi_i; y_i) &=& \begin{cases}
 -\tau_i- \left(1+\cfrac{1}{\xi_i} \right) \log(1+z_i) - (1+z_i)^{-1/\xi_i}, & y_i \in\Omega, \\
 -\tau_i-\exp(-z_i)-z_i, & (y_i, \xi_i) \in \Omega_0,
\end{cases} 
\end{eqnarray*}
with 
$$
z_i =
\begin{cases}
(y_i-\mu_i)\xi_i\exp(-\tau_i), & y_i \in\Omega,\\
(y_i-\mu_i)\exp(-\tau_i), & (y_i, \xi_i) \in \Omega_0.
\end{cases}
$$ 
This log-likelihood becomes numerically unstable when $\xi_i$ and $z_i$ are close to zero, while overflow is amplified as the order of the derivatives increases. The proposed approximate EM method requires third-order log-likelihood derivatives, which involve terms like $\xi_i^{-5}$. When $\xi_i\approx0$, the threshold {below} which the absolute value of the shape parameter should be set to zero is therefore troublesome. Its value should reflect the compromise between stability of the derivatives and the switch from the general GEV form to the Gumbel distribution. The numerical instability is even more problematic in the \verb|mgcv| gam method, which requires fourth-order log-likelihood derivatives; the lower the order of the derivatives, the fewer unstable computations. In our implementation, we set $\xi_i=0$ whenever $|\xi_i| \leq \varpi^{3/10}$, with $\varpi$  the machine precision. This sets the order of the threshold to $10^{-5}$, while allowing negative exponents of $\xi_i$ terms to grow up to $10^{15}$, which is within the range of precision of all modern machines.

\begin{figure}[t]
\vskip 0.2in
\begin{center}
\centerline{\includegraphics[width=0.55\textheight]{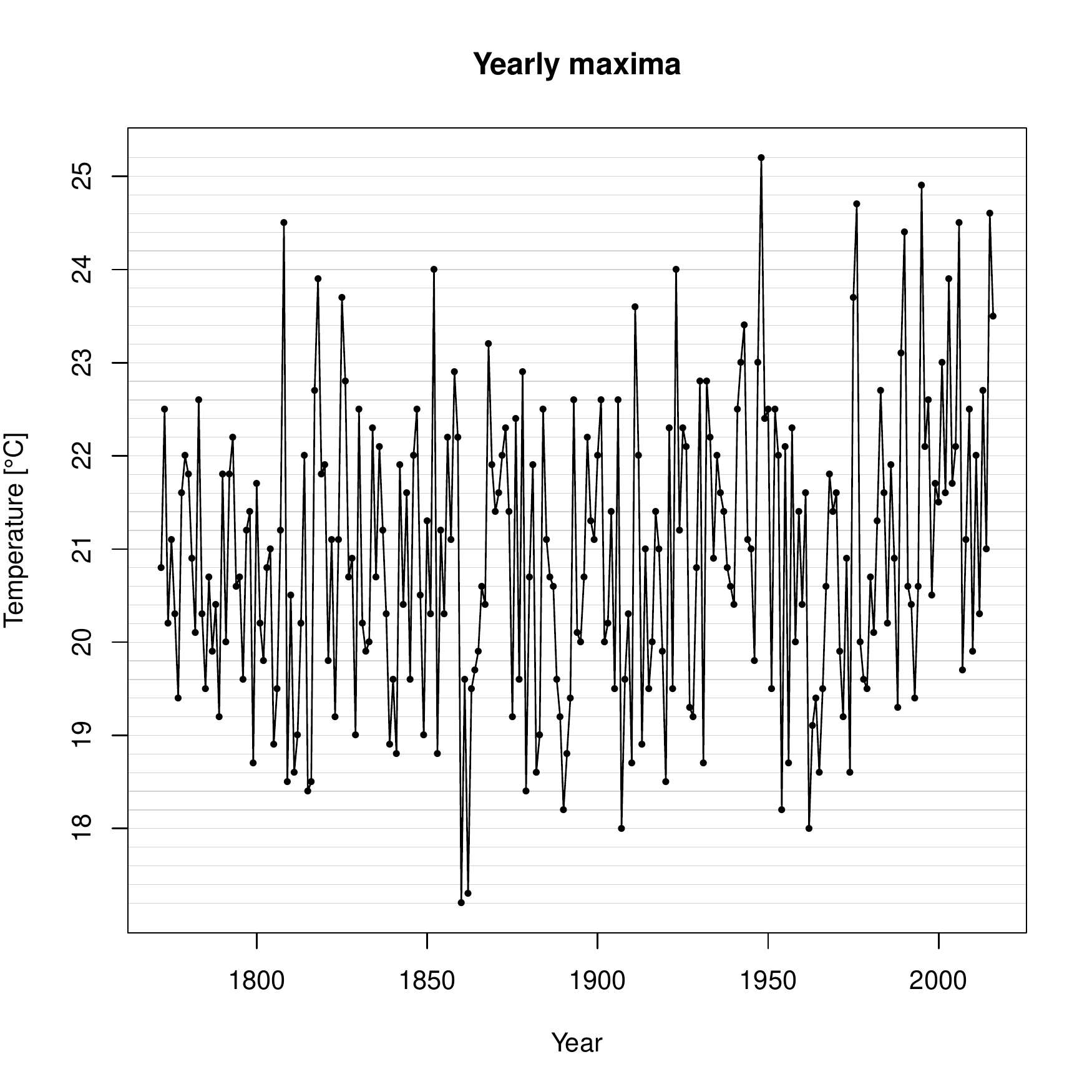}}
\caption{Yearly maxima taken over the months.}
\label{fig:appliexplo1}
\end{center}
\vskip -0.2in
\end{figure}

\subsection{Application}\label{sec:application}
\begin{figure}[t]
  \centering
  \begin{minipage}[b]{0.4\textwidth}
    \centerline{\includegraphics[width=0.42\textheight]{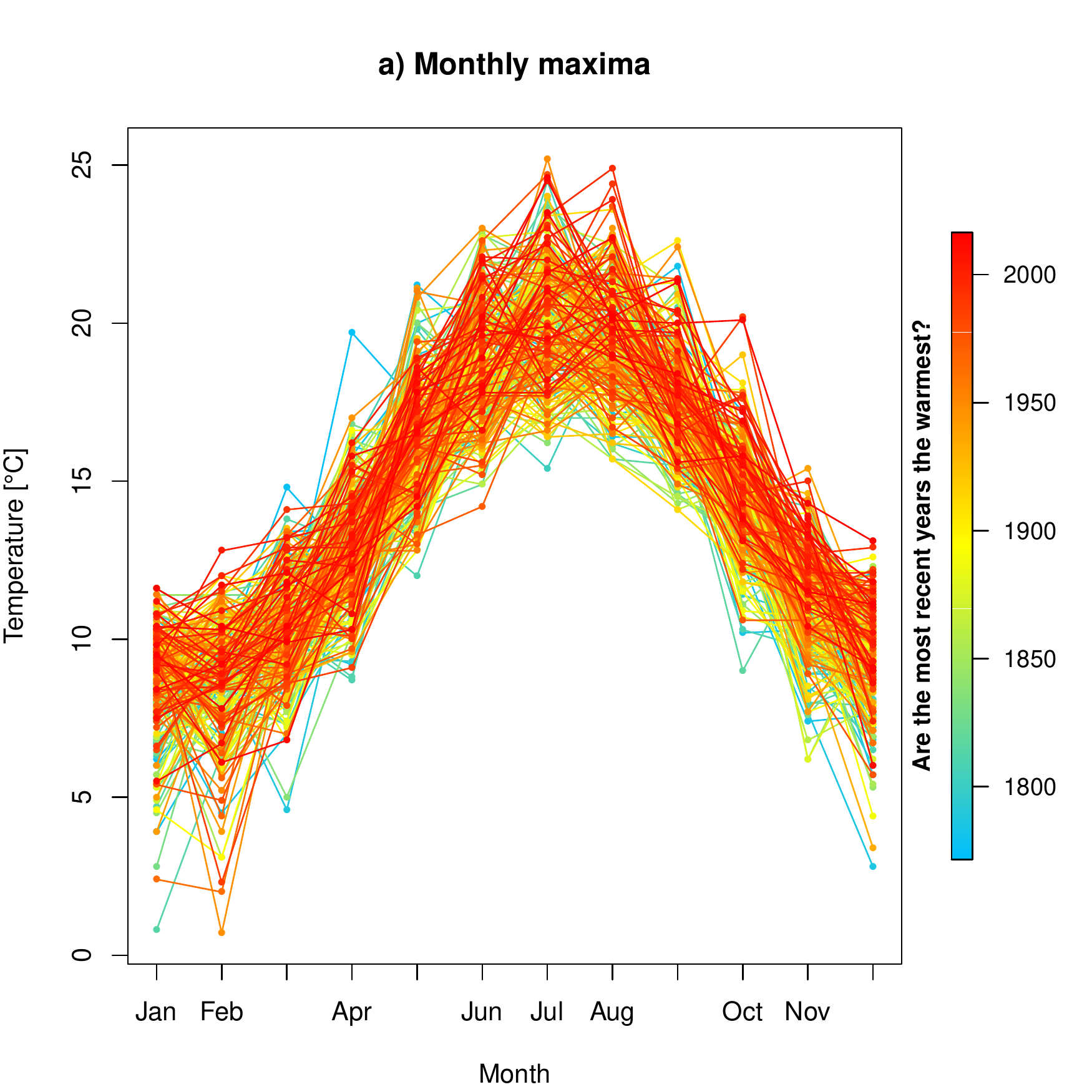}}
  \end{minipage}
  \hfill
  \begin{minipage}[b]{0.43\textwidth}
    \centerline{\includegraphics[width=0.42\textheight]{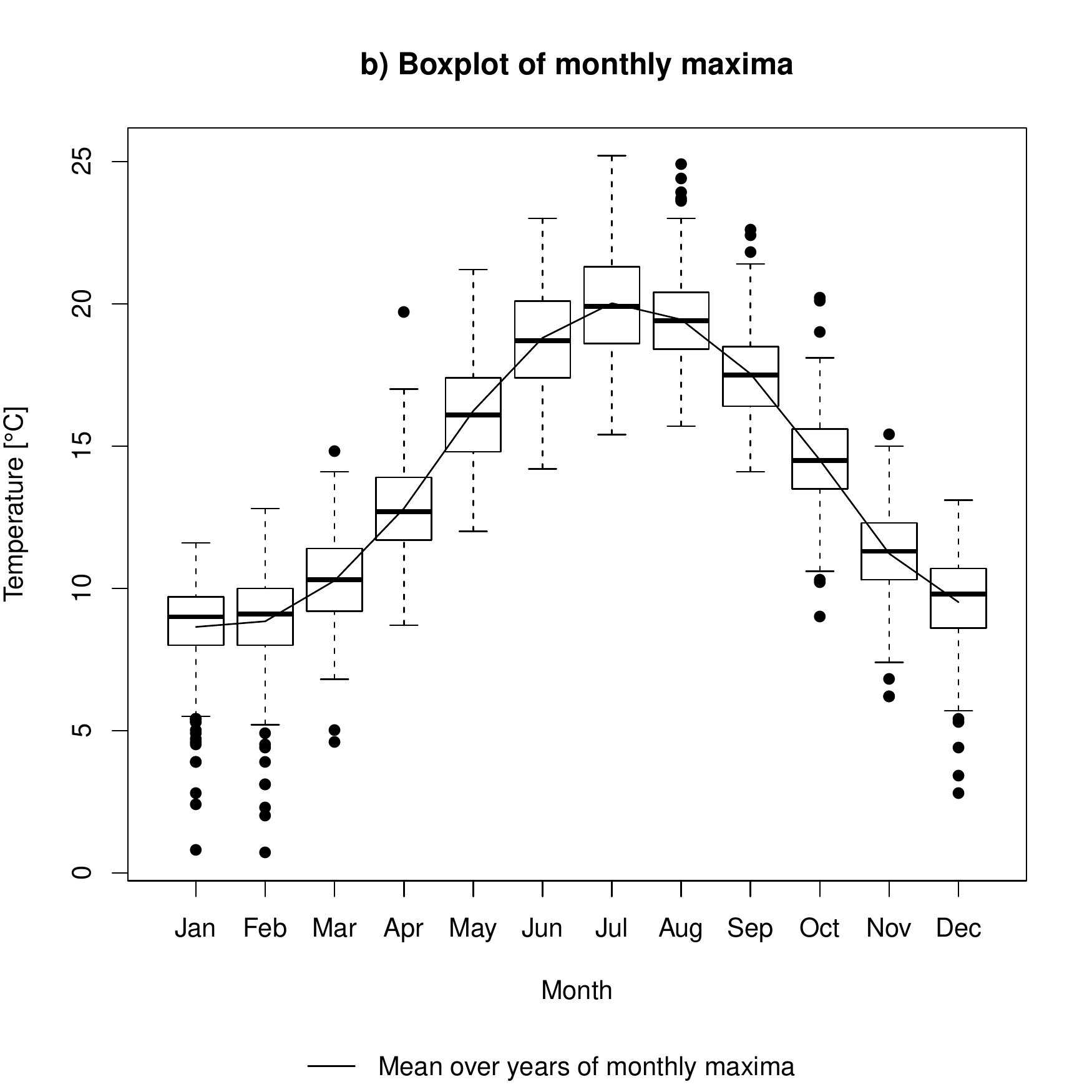}}
  \end{minipage}
  \caption{Monthly maxima.}
  \label{fig:appliexplo2}
\end{figure}

We analyze monthly maxima of the daily Central England Temperature (CET){\footnote{\url{https://www.metoffice.gov.uk/hadobs/hadcet/data/download.html}}} series from January 1772 to December 2016. Figure~\ref{fig:appliexplo1} shows yearly maxima and suggests that the recent years are the warmest, while panel a) in Figure~\ref{fig:appliexplo2} indicates that any increase is most apparent at the end of the year. Figure~\ref{fig:appliexplo2} exhibits obvious seasonality, which we represent using 12 basis functions from cyclic cubic regression splines for each of the location, scale and shape parameters of the GEV model; we use ten basis functions from thin plate splines~\citep{Wood2003} in the location for the trend visible in Figure~\ref{fig:appliexplo1}. We included trend in the scale and shape initially, but these were not significant. To our knowledge, this is the only paper modeling a variable shape parameter for this dataset. Neither of the algorithms in Stan~\citep{stan} using the \verb|R| package \verb|brms| \citep{brms} converged and the variational Bayes approach faced numerical instabilities. 

\begin{figure}[!t]
  \centering
  \begin{minipage}[b]{0.4\textwidth}
    \centerline{\includegraphics[width=0.42\textheight]{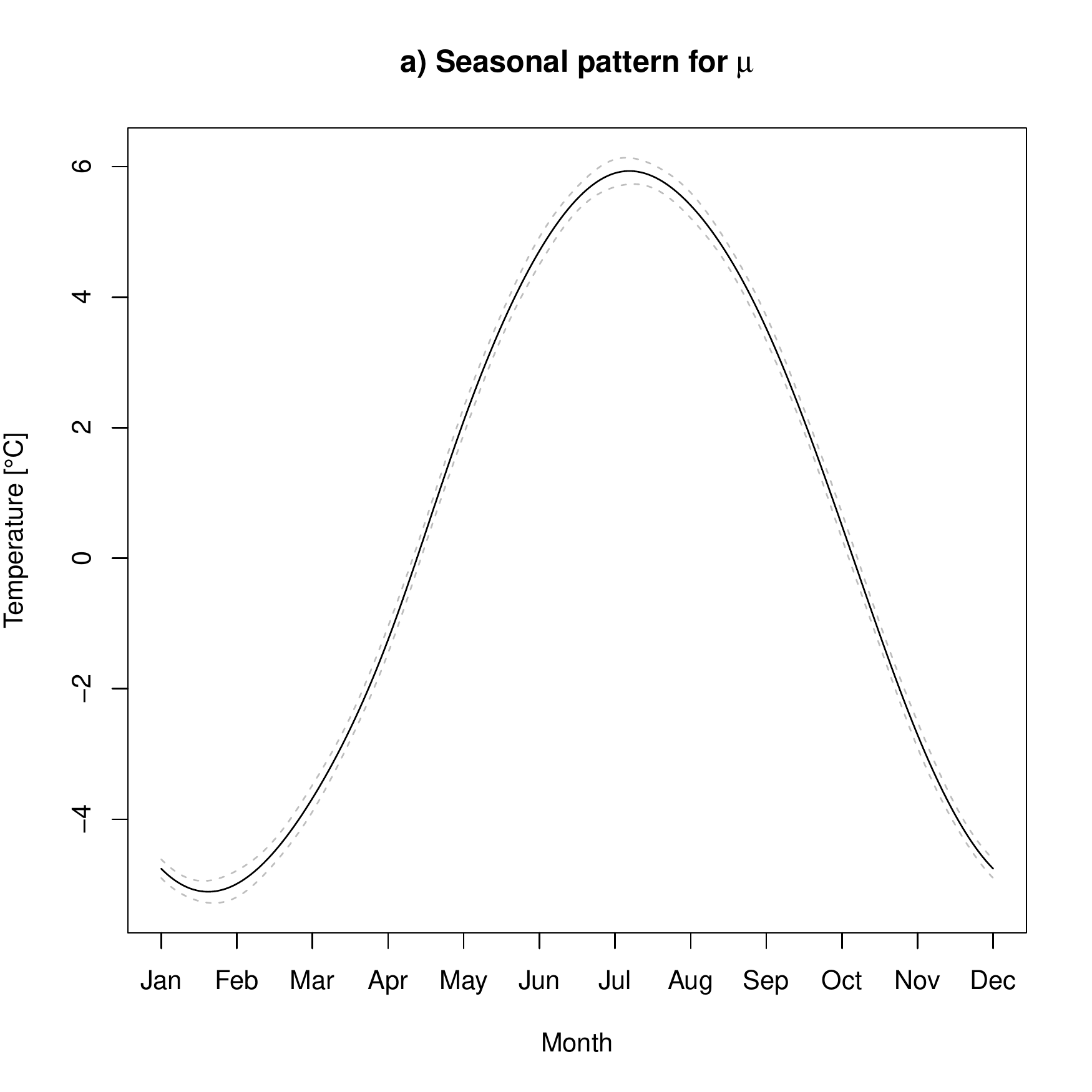}}
  \end{minipage}
  \hfill
  \begin{minipage}[b]{0.43\textwidth}
   \centerline{\includegraphics[width=0.42\textheight]{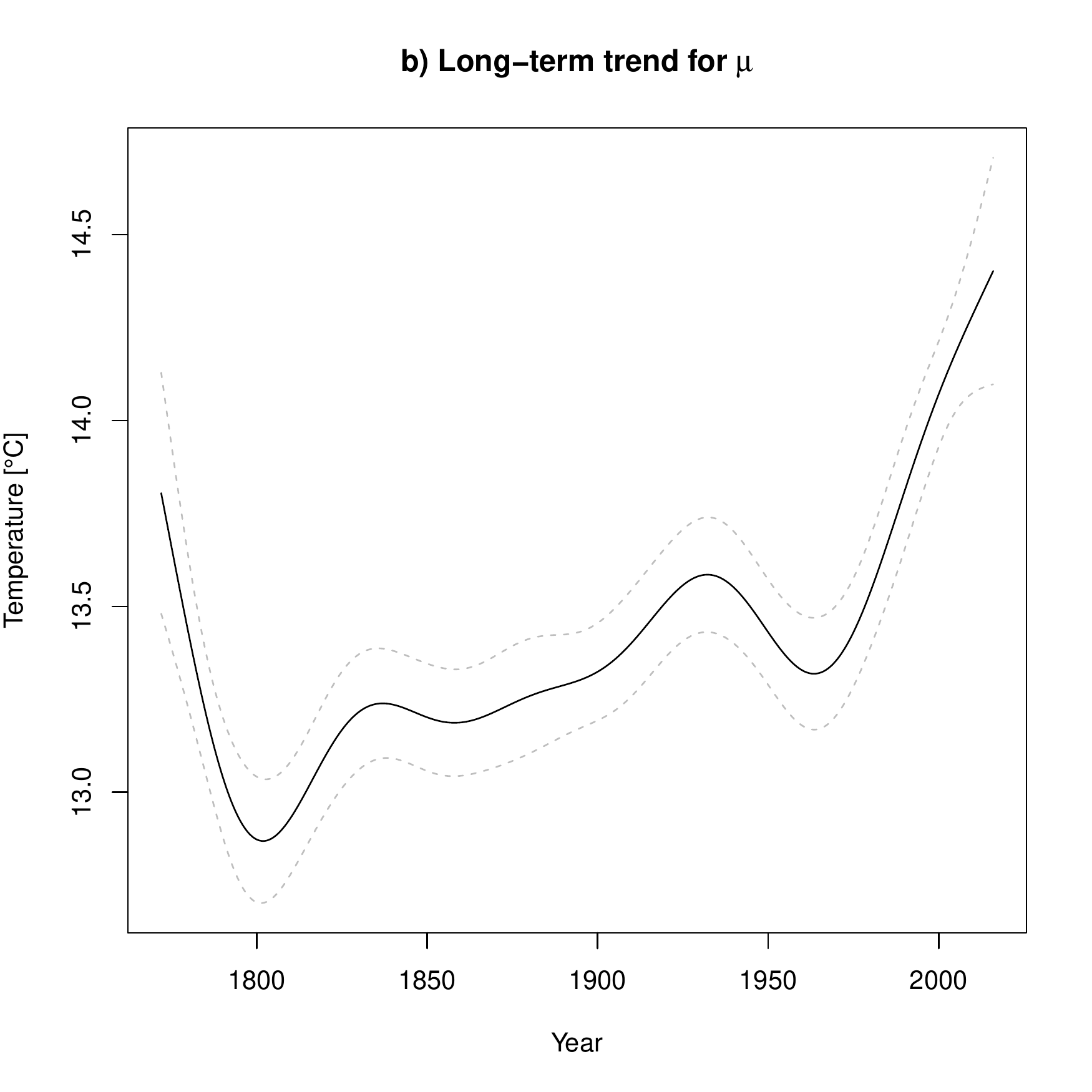}}
  \end{minipage}
  \centering
  \begin{minipage}[b]{0.4\textwidth}
    \centerline{\includegraphics[width=0.43\textheight]{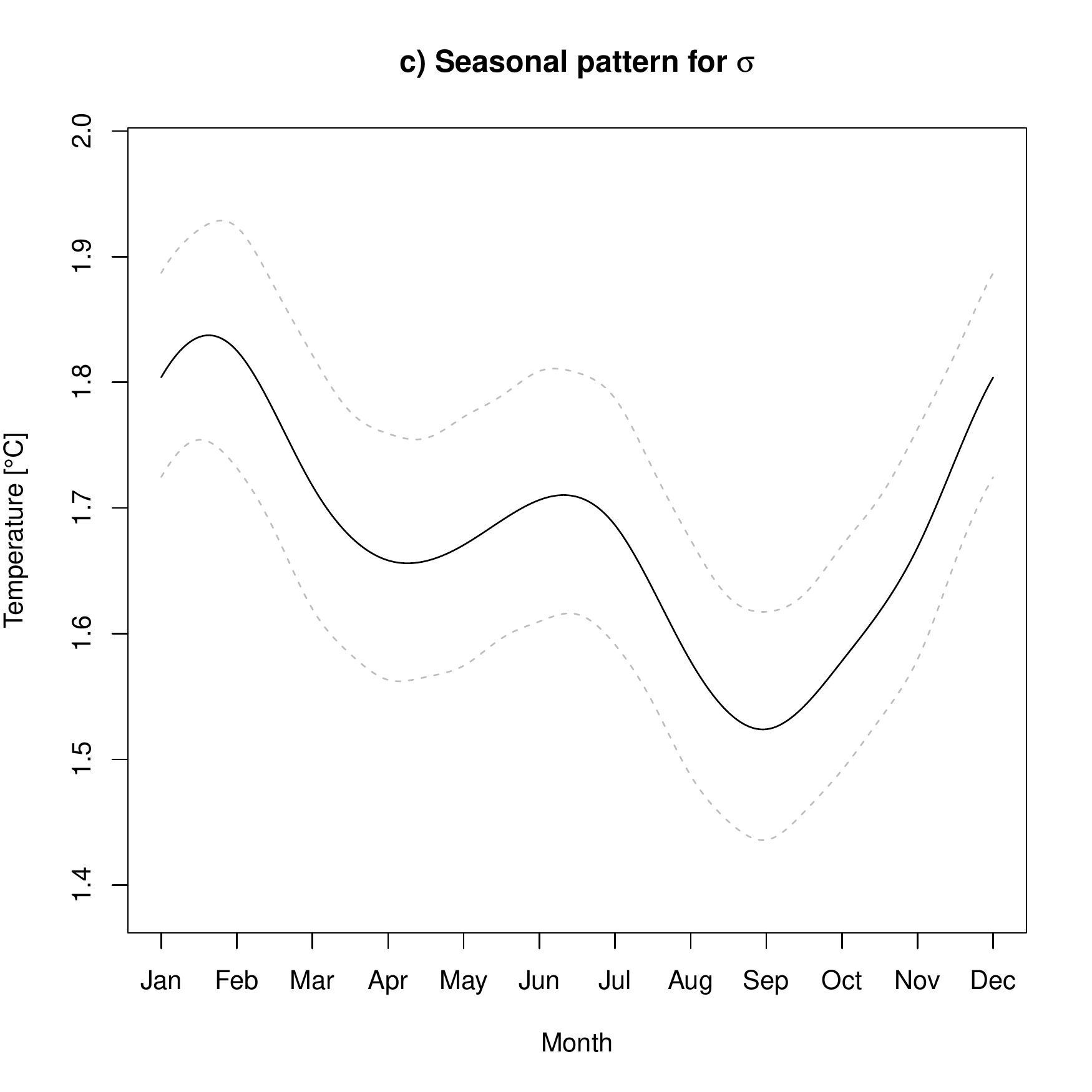}}
  \end{minipage}
  \hfill
  \begin{minipage}[b]{0.43\textwidth}
    \centerline{\includegraphics[width=0.43\textheight]{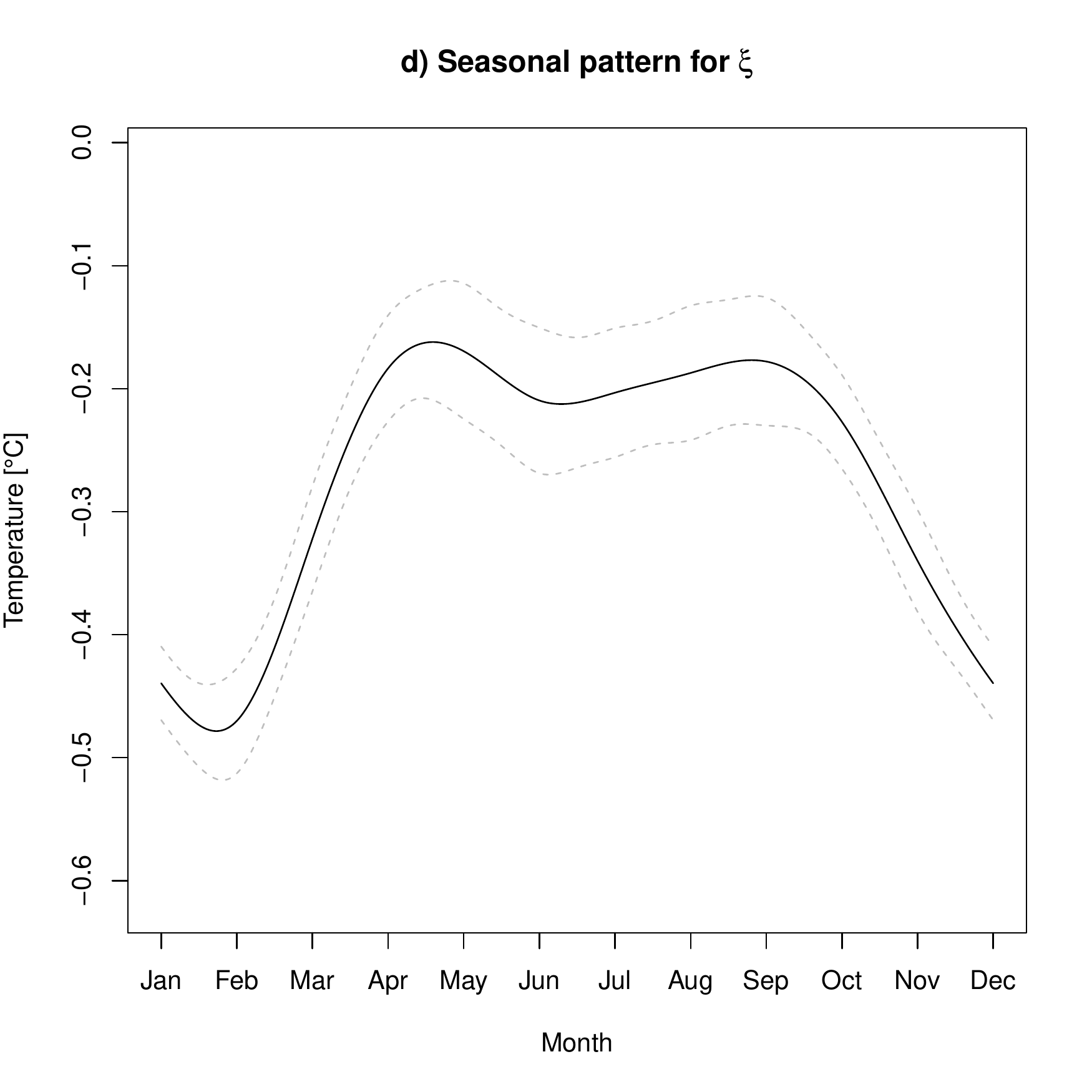}}
  \end{minipage}
  \caption{Learned functional parameters, with $95\%$ pointwise confidence intervals (dashes).}
  \label{fig:result2}
\end{figure}

Panel a) of Figure~\ref{fig:result2} shows an annual change of $11\degree $C similar to that in the empirical version in Figure~\ref{fig:appliexplo2}, and panel b) of Figure~\ref{fig:result2} illustrates a non-linear trend with a drop from 1772 to 1800 and a sharp increase from the 1960s onwards. The pattern between is hard to discern in Figure~\ref{fig:appliexplo1}, but panel b) shows an overall increase of about $1.5\degree $C from 1800 onwards and peaks over the last few decades. The learned scale and shape parameters in Figure~\ref{fig:result2}, whose functional forms vary significantly through the year, give insight into the seasonality. They are negatively correlated except in mid-June to September, where the increase in the shape is much slower and weaker than the drop in the scale. We can distinguish two cycles within the year, with similar patterns but different intensities: the extended strong winter from September to April, and the extended weak summer, from April to September. Each of these incorporates two antagonistic phases which are negatively correlated, alternating between decrease and increase for the shape, and vice-versa for the scale. Figure~\ref{fig:result2} summarizes the influences of the scale and the shape parameters on the seasonality of the CET data as follows: whether the temperature is increasing or decreasing seems to be smoothly related to the direction of the shape in the winter, and to that of the scale in the summer. Since the former controls the tail of the distribution and is always significantly negative here, the temperature is bounded above throughout the year; the strongest increase of the shape occurs in February to mid-April, early spring, stabilizing around its highest values, $-0.2$ or so, in the summer. This stabilization and the negative correlation between the scale and the shape explain why the sharper fluctuations of the scale have more impact on the temperature in the summer than the near-constant shape. The rather narrow pointwise confidence intervals suggest that there is very strong evidence for seasonal variation of the shape, and less strong but still appreciable evidence of such variation for the scale. 

\begin{figure}[!t]
\vskip 0.2in
\begin{center}
\centerline{\includegraphics[width=0.45\textheight]{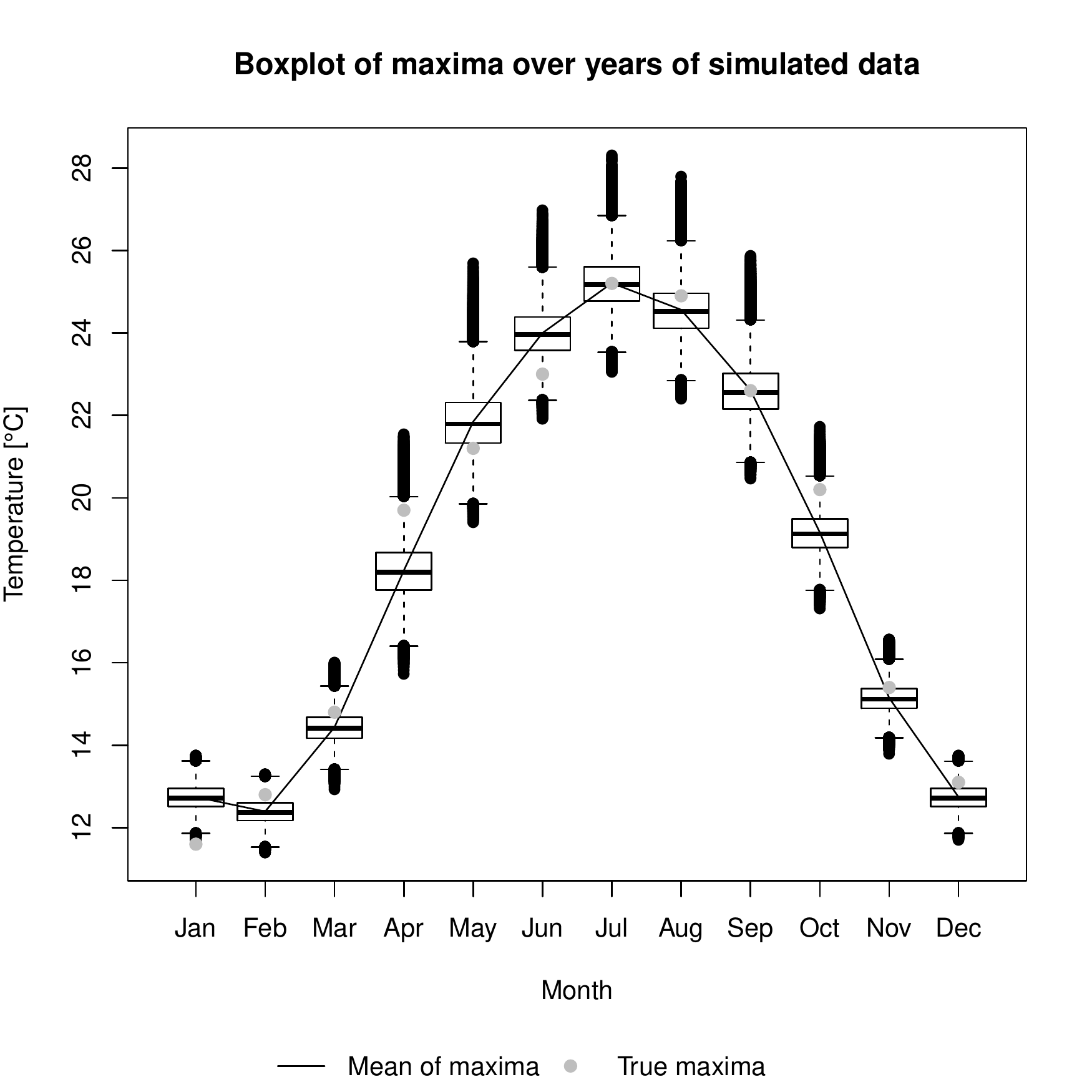}}
\caption{Monthly maxima simulated from the learned GEV model.}
\label{fig:result5}
\end{center}
\vskip -0.2in
\end{figure}

\begin{figure}[!t]
  \centering
  \begin{minipage}[b]{0.4\textwidth}
    \centerline{\includegraphics[width=0.43\textheight]{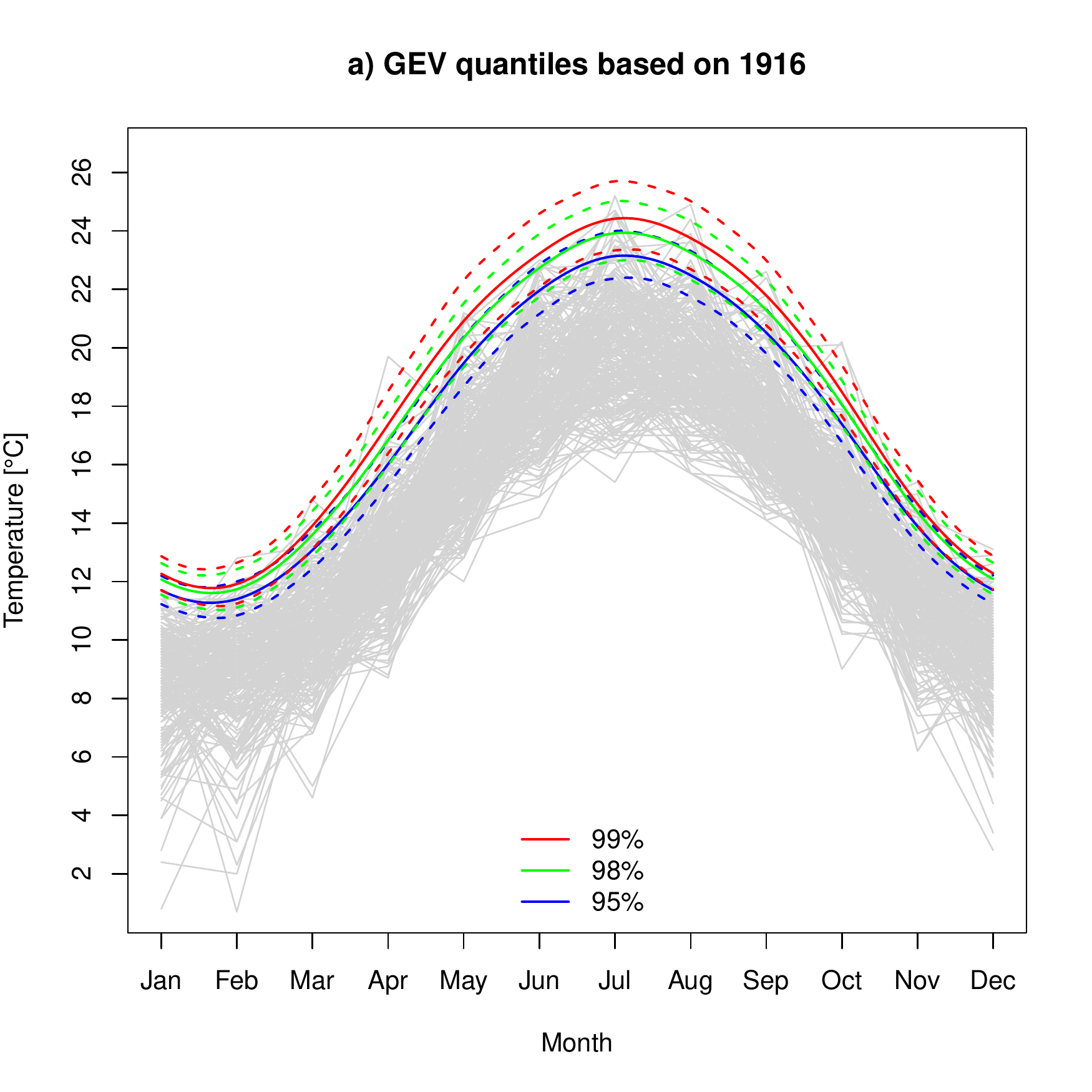}}
  \end{minipage}
  \hfill
  \begin{minipage}[b]{0.43\textwidth}
    \centerline{\includegraphics[width=0.43\textheight]{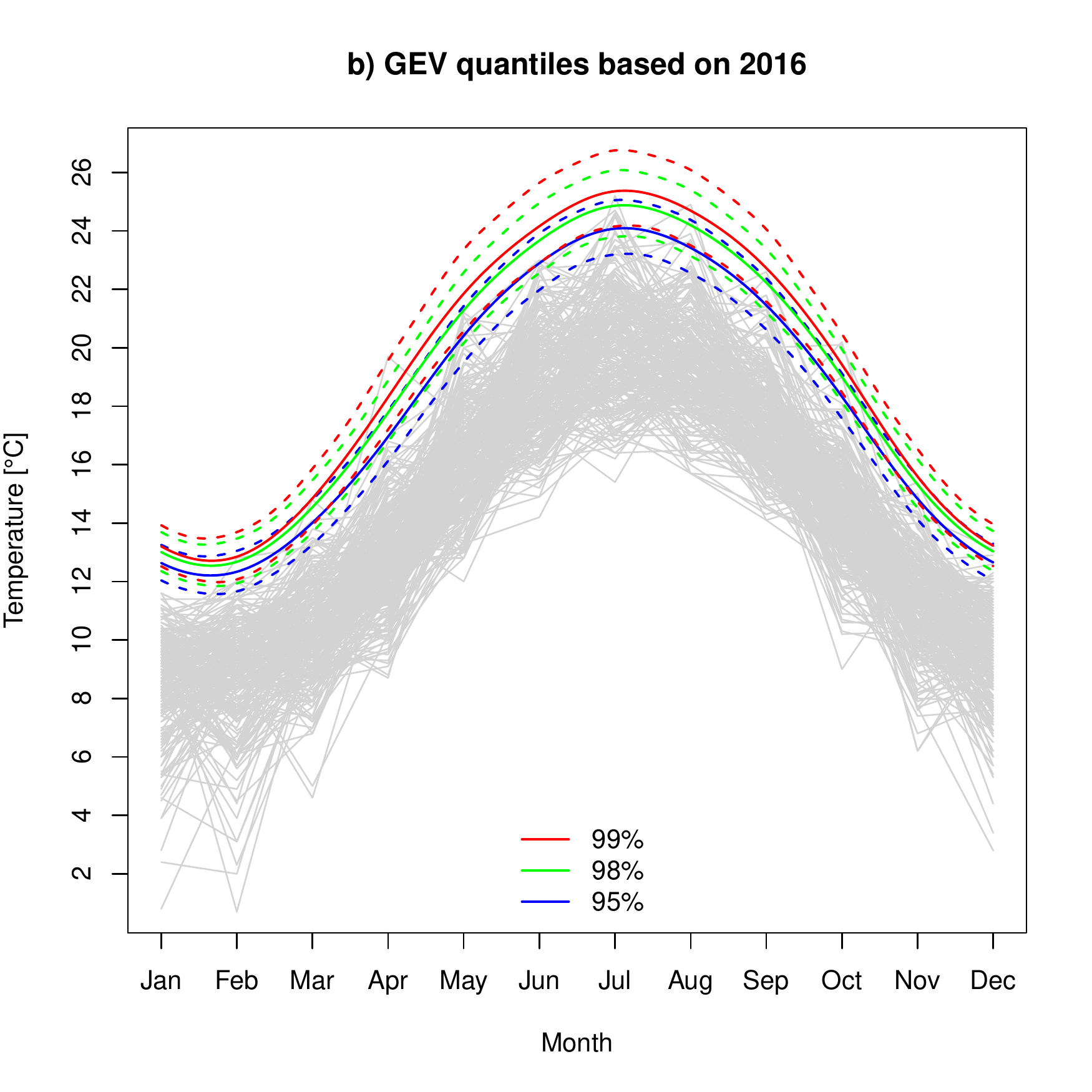}}
  \end{minipage}
  \caption{Superposition of the original data (grey) and quantiles of the GEV models, with pointwise confidence intervals (dashes).}
  \label{fig:result6}
\end{figure}

Figures~\ref{fig:result5} and~\ref{fig:result6} illustrate diagnostics of model fit. Figure~\ref{fig:result5} shows that the true maxima are within the range of those simulated from the learned model. Figure~\ref{fig:result6} represents the predicted 0.95, 0.98 and 0.99 quantiles for monthly maxima. Based on the model for 1916, only one value from previous years, $24.5\degree $C in July 1808, exceeded the maximum of the 0.99 quantile curve, $24.4\degree $C, in July; all other exceedances occur after 1916. The maximum of the 0.99 quantile curve in 2016 occurred in July at $25.4\degree $C, and no higher temperature has been observed. Overall, the model does not seem unrealistic, although it may underestimate slightly the uncertainty, as it assumes independence of maxima in successive months. A possible improvement would be a GEV model with multiple GAMs and autoregressive errors.



\section{Discussion}\label{sec:discussion}
This paper makes contributions to optimal smoothing for multiple generalized additive models, where the smoothing penalty corresponds to a weighted $L_2$ regularization that is interpreted as a Gaussian prior on the regression weights, and whose posterior is the penalized log-likelihood. We adopt an empirical Bayes approach for optimizing the log-marginal likelihood to obtain the appropriate smoothing parameters automatically. This uses an EM algorithm which is made tractable using a double Laplace approximation of the moment generating function underlying the E-step. The new approach transfers maximization of the log-marginal likelihood to a function whose maximizer has a closed form, and avoids evaluation of expensive and numerically unstable terms. The only requirement is that the log-likelihood has third derivatives. The new method is stable, accurate and fast. Its stability is ensured both by the EM approach and by its need for fewer derivatives, making the proposed method broadly applicable for complex models. Its high accuracy is established theoretically by~\citet{TierneyKassKadane1989}, with an $O(n^{-2})$ error in the E-step approximation. Its serial implementation is substantially faster than the best existing methods and achieves state-of-the-art accuracy. It can easily be parallelized, making it appealing for extension to big-data settings, where no reliable method yet exists.

These advantages are balanced by potential difficulties. First, the EM algorithm can be slow around the optimum. Tests show that this happens when certain smoothing parameters become so large that their corresponding smooth functions are linear, and their updates no longer change the penalized log-likelihood. At that point, we declare convergence for those components of $\boldsymbol{\lambda}$, though they may keep changing without affecting the regression weights. Validating convergence for a portion of smoothing parameters and updating the remainder is supported by the diagonality of the Hessian matrix at the E-step. Second, the EM is known to suffer from local optima, though we found none in the datasets and the simulated models we analyzed, perhaps because the log-likelihood is fairly quadratic for large samples. 

The proposed method is implemented in a C++ library that uses Eigen \citep{eigen} for matrix decompositions, is integrated into the \verb|R| package \verb|multgam| through the interface~\verb|RcppEigen| \citep{rcppeigen}, and makes addition of further probability models straightforward.





\vskip 0.2in
\bibliography{references.bib}

\begin{thebibliography}{36}
\providecommand{\natexlab}[1]{#1}
\providecommand{\url}[1]{\texttt{#1}}
\expandafter\ifx\csname urlstyle\endcsname\relax
  \providecommand{\doi}[1]{doi: #1}\else
  \providecommand{\doi}{doi: \begingroup \urlstyle{rm}\Url}\fi

\bibitem[Bates and Eddelbuettel(2013)]{rcppeigen}
D.~Bates and D.~Eddelbuettel.
\newblock Fast and {E}legant {N}umerical {L}inear {A}lgebra {U}sing the
  {RcppEigen} {P}ackage.
\newblock \emph{Journal of Statistical Software}, 52\penalty0 (5):\penalty0
  1--24, 2013.
\newblock URL \url{http://www.jstatsoft.org/v52/i05/}.

\bibitem[Breiman and Friedman(1985)]{BreimanFriedman}
L.~Breiman and J.~H. Friedman.
\newblock Estimating {O}ptimal {T}ransformations for {M}ultiple {R}egression
  and {C}orrelation.
\newblock \emph{Journal of the American Statistical Association}, 80\penalty0
  (391):\penalty0 580--598, 1985.
\newblock ISSN 01621459.
\newblock URL \url{http://www.jstor.org/stable/2288473}.

\bibitem[Burkner(2017)]{brms}
J.~C. Burkner.
\newblock {brms}: An {R} package for {Bayesian} {M}ultilevel {M}odels {U}sing
  {S}tan.
\newblock \emph{Journal of Statistical Software}, 80\penalty0 (1):\penalty0
  1--28, 2017.

\bibitem[Carpenter et~al.(2017)Carpenter, Lee, Brubaker, Riddell, Gelman,
  Goodrich, Guo, Hoffman, Betancourt, and Li]{stan}
B.~Carpenter, D.~Lee, M.~A. Brubaker, A.~Riddell, A.~Gelman, B.~Goodrich,
  J.~Guo, M.~Hoffman, M.~Betancourt, and P.~Li.
\newblock Stan: {A} {P}robabilistic {P}rogramming {L}anguage, 2017.

\bibitem[Chavez-Demoulin and Davison(2005)]{Chavez2005}
V.~Chavez-Demoulin and A.~C. Davison.
\newblock Generalized {A}dditive {M}odelling of {S}ample {E}xtremes.
\newblock \emph{Journal of the Royal Statistical Society, Series C},
  54\penalty0 (1):\penalty0 207--222, 2005.
\newblock ISSN 00359254, 14679876.
\newblock URL \url{http://www.jstor.org/stable/3592608}.

\bibitem[Chavez-Demoulin and Davison(2012)]{chavez2012}
V.~Chavez-Demoulin and A.~C. Davison.
\newblock Modelling {T}ime {S}eries {E}xtremes.
\newblock \emph{Revstat-Statistical Journal}, 10:\penalty0 109--133, 2012.

\bibitem[Cleveland et~al.(1993)Cleveland, Grosse, and Shyu]{Loess1993}
W.~S. Cleveland, E.~Grosse, and W.~M. Shyu.
\newblock \emph{Local {R}egression {M}odels.}
\newblock Chapman \& Hall, New York, 1993.

\bibitem[Cole and Green(1992)]{ColesGreen1992}
T.~J. Cole and P.~J. Green.
\newblock Smoothing {R}eference {C}entile {C}urves: the {L}{M}{S} {M}ethod and
  {P}enalized {L}ikelihood.
\newblock \emph{Statistics in Medicine}, 11\penalty0 (10):\penalty0 1305--1319,
  1992.
\newblock ISSN 1097-0258.
\newblock \doi{10.1002/sim.4780111005}.
\newblock URL \url{http://dx.doi.org/10.1002/sim.4780111005}.

\bibitem[de~Haan and Ferreira(2006)]{HaanFerreira}
L.~de~Haan and A.~Ferreira.
\newblock \emph{Extreme Value Theory}.
\newblock Springer-Verlag New York, 2006.
\newblock ISBN 978-0-387-23946-0.

\bibitem[Dempster et~al.(1977)Dempster, Laird, and Rubin]{EM77}
A.~P. Dempster, N.~M. Laird, and D.~B. Rubin.
\newblock Maximum {L}ikelihood from {I}ncomplete {D}ata via the {E}{M}
  {A}lgorithm (with {D}iscussion).
\newblock \emph{Journal of the Royal Statistical Society, Series B},
  39\penalty0 (1):\penalty0 1--38, 1977.

\bibitem[Fisher and Tippett(1928)]{FisherTippett1928}
R.~A. Fisher and L.~H.~C. Tippett.
\newblock Limiting {F}orms of the {F}requency {D}istributions of the {L}argest
  or {S}mallest {M}ember of a {S}ample.
\newblock \emph{Proceedings of the Cambridge Philosophical Society},
  24:\penalty0 180--190, 1928.

\bibitem[Golub and Van~Loan(2013)]{golub}
G.~H. Golub and C.~F. Van~Loan.
\newblock \emph{Matrix Computations}.
\newblock The Johns Hopkins University Press, Baltimore, Maryland, 4 edition,
  2013.

\bibitem[Gu(1992)]{Gu1992}
C.~Gu.
\newblock Cross-{V}alidating {N}on-{G}aussian {D}ata.
\newblock \emph{Journal of Computational and Graphical Statistics}, 1\penalty0
  (2):\penalty0 169--179, 1992.
\newblock \doi{10.1080/10618600.1992.10477012}.

\bibitem[Guennebaud et~al.(2018)Guennebaud, Jacob, et~al.]{eigen}
G.~Guennebaud, B.~Jacob, et~al.
\newblock Eigen v3.
\newblock http://eigen.tuxfamily.org, 2018.

\bibitem[Hastie and Tibshirani(1986)]{hastie1986}
T.~J. Hastie and R.~J. Tibshirani.
\newblock Generalized {A}dditive {M}odels (with {D}iscussion).
\newblock \emph{Statistical Science}, 1:\penalty0 297--310, 1986.

\bibitem[Hastie and Tibshirani(1990)]{HastieTib}
T.~J. Hastie and R.~J. Tibshirani.
\newblock \emph{Generalized Additive Models}.
\newblock Chapman \& Hall, 1990.

\bibitem[Jenkinson(1955)]{Jenkinson1955}
A.~F. Jenkinson.
\newblock The {F}requency {D}istribution of the {A}nnual {M}aximum (or
  {M}inimum) {V}alues of {M}eteorological {E}lements.
\newblock \emph{Journal of the Royal Meteorological Society}, 81:\penalty0
  158--171, 1955.

\bibitem[Kimeldorf and Wahba(1970)]{kimeldorf1970}
G.~S. Kimeldorf and G.~Wahba.
\newblock A {C}orrespondence {B}etween {B}ayesian {E}stimation on {S}tochastic
  {P}rocesses and {S}moothing by {S}plines.
\newblock \emph{The Annals of Mathematical Statistics}, 41\penalty0
  (2):\penalty0 495--502, 1970.
\newblock URL \url{http://dx.doi.org/10.1214/aoms/1177697089}.

\bibitem[McLachlan and Krishnan(2008)]{EMbook2008}
G.~J. McLachlan and T.~Krishnan.
\newblock \emph{The EM Algorithm and Extensions (Wiley Series in Probability
  and Statistics)}.
\newblock Wiley-Interscience, 2 edition, 2008.
\newblock ISBN 0471201707.

\bibitem[Nelder and Wedderburn(1972)]{nelderwedderburn1972}
J.~A. Nelder and R.~W.~M. Wedderburn.
\newblock Generalized {L}inear {M}odels.
\newblock \emph{Journal of the Royal Statistical Society, Series A},
  135\penalty0 (3):\penalty0 370--384, 1972.
\newblock ISSN 00359238.
\newblock URL \url{http://www.jstor.org/stable/2344614}.

\bibitem[Oakes(1999)]{Oakes1999}
D.~Oakes.
\newblock Direct {C}alculation of the {I}nformation {M}atrix via the {E}{M}.
\newblock \emph{Journal of the Royal Statistical Society, Series B},
  61\penalty0 (2):\penalty0 479--482, 1999.
\newblock ISSN 1467-9868.
\newblock \doi{10.1111/1467-9868.00188}.
\newblock URL \url{http://dx.doi.org/10.1111/1467-9868.00188}.

\bibitem[O'Sullivan et~al.(1986)O'Sullivan, Yandell, and Raynor]{OSullivan1986}
F.~O'Sullivan, B.~S. Yandell, and W.~J. Raynor.
\newblock Automatic {S}moothing of {R}egression {F}unctions in {G}eneralized
  {L}inear {M}odels.
\newblock \emph{Journal of the American Statistical Association}, 81\penalty0
  (393):\penalty0 96--103, 1986.

\bibitem[{R Core Team}(2018)]{RR}
{R Core Team}.
\newblock \emph{R: A {L}anguage and {E}nvironment for {S}tatistical
  {C}omputing}.
\newblock R Foundation for Statistical Computing, Vienna, Austria, 2018.
\newblock URL \url{https://www.R-project.org/}.

\bibitem[Reiss and Ogden(2009)]{ReissOgden}
P.~T. Reiss and R.~T. Ogden.
\newblock Smoothing {P}arameter {S}election for a {C}lass of {S}emiparametric
  {L}inear {M}odels.
\newblock \emph{Journal of the Royal Statistical Society, Series B},
  71\penalty0 (2):\penalty0 505--523, 2009.
\newblock URL
  \url{http://EconPapers.repec.org/RePEc:bla:jorssb:v:71:y:2009:i:2:p:505-523}.

\bibitem[Rigby and Stasinopoulos(1996)]{Rigby1996a}
R.~A. Rigby and D.~M. Stasinopoulos.
\newblock A {S}emi-parametric {A}dditive {M}odel for {V}ariance
  {H}eterogeneity.
\newblock \emph{Statistics and Computing}, 6\penalty0 (1):\penalty0 57--65,
  1996.
\newblock ISSN 1573-1375.
\newblock \doi{10.1007/BF00161574}.
\newblock URL \url{http://dx.doi.org/10.1007/BF00161574}.

\bibitem[Rigby and Stasinopoulos(2005)]{RigbyStasinopoulos2005}
R.~A. Rigby and D.~M. Stasinopoulos.
\newblock Generalized {A}dditive {M}odels for {L}ocation, {S}cale and {S}hape
  (with discussion).
\newblock \emph{Journal of the Royal Statistical Society, Series C},
  54\penalty0 (3):\penalty0 507--554, 2005.
\newblock ISSN 1467-9876.
\newblock \doi{10.1111/j.1467-9876.2005.00510.x}.
\newblock URL \url{http://dx.doi.org/10.1111/j.1467-9876.2005.00510.x}.

\bibitem[Rue et~al.(2009)Rue, Martino, and Chopin]{inla}
H.~Rue, S.~Martino, and N.~Chopin.
\newblock Approximate {B}ayesian {I}nference for {L}atent {G}aussian {M}odels
  by {U}sing {I}ntegrated {N}ested {L}aplace {A}pproximations.
\newblock \emph{Journal of the Royal Statistical Society: Series B (Statistical
  Methodology)}, 71\penalty0 (2):\penalty0 319--392, 2009.

\bibitem[Silverman(1985)]{Silverman1985}
B.~W. Silverman.
\newblock Some {A}spects of the {S}pline {S}moothing {A}pproach to
  {N}on-{P}arametric {R}egression {C}urve {F}itting.
\newblock \emph{Journal of the Royal Statistical Society, Series B},
  47\penalty0 (1):\penalty0 1--52, 1985.
\newblock ISSN 00359246.
\newblock URL \url{http://www.jstor.org/stable/2345542}.

\bibitem[Tierney et~al.(1989)Tierney, Kass, and Kadane]{TierneyKassKadane1989}
L.~Tierney, R.~E. Kass, and J.~B. Kadane.
\newblock Fully {E}xponential {L}aplace {A}pproximations to {E}xpectations and
  {V}ariances of {N}onpositive {F}unctions.
\newblock \emph{Journal of the American Statistical Association}, 84\penalty0
  (407):\penalty0 710--716, 1989.
\newblock \doi{10.1080/01621459.1989.10478824}.

\bibitem[Wood(2003)]{Wood2003}
S.~N. Wood.
\newblock Thin {P}late {R}egression {S}plines.
\newblock \emph{Journal of the Royal Statistical Society, Series B},
  65\penalty0 (1):\penalty0 95--114, 2003.
\newblock ISSN 1467-9868.
\newblock \doi{10.1111/1467-9868.00374}.
\newblock URL \url{http://dx.doi.org/10.1111/1467-9868.00374}.

\bibitem[Wood(2008)]{Wood2008}
S.~N. Wood.
\newblock Fast {S}table {D}irect {F}itting and {S}moothness {S}election for
  {G}eneralized {A}dditive {M}odels.
\newblock \emph{Journal of the Royal Statistical Society, Series B},
  70\penalty0 (3):\penalty0 495--518, 2008.
\newblock \doi{10.1111/j.1467-9868.2007.00646.x}.
\newblock URL \url{http://opus.bath.ac.uk/16622/}.

\bibitem[Wood(2011)]{Wood2011}
S.~N. Wood.
\newblock Fast {S}table {R}estricted {M}aximum {L}ikelihood and {M}arginal
  {L}ikelihood {E}stimation of {S}emiparametric {G}eneralized {L}inear
  {M}odels.
\newblock \emph{Journal of the Royal Statistical Society, Series B},
  73\penalty0 (1):\penalty0 3--36, 2011.
\newblock \doi{10.1111/j.1467-9868.2010.00749.x}.
\newblock URL \url{http://opus.bath.ac.uk/22707/}.

\bibitem[Wood et~al.(2015)Wood, Goude, and Shaw]{Wood2015}
S.~N. Wood, Y.~Goude, and S.~Shaw.
\newblock Generalized {A}dditive {M}odels for {L}arge {D}ata {S}ets.
\newblock \emph{Journal of the Royal Statistical Society: Series C (Applied
  Statistics)}, 64\penalty0 (1):\penalty0 139--155, 2015.

\bibitem[Wood et~al.(2016)Wood, Pya, and Safken]{Wood2016}
S.~N. Wood, N.~Pya, and B.~Safken.
\newblock Smoothing {P}arameter and {M}odel {S}election for {G}eneral {S}mooth
  {M}odels.
\newblock \emph{Journal of the American Statistical Association}, 111\penalty0
  (516):\penalty0 1548--1563, 2016.
\newblock \doi{10.1080/01621459.2016.1180986}.
\newblock URL \url{http://dx.doi.org/10.1080/01621459.2016.1180986}.

\bibitem[Wood et~al.(2017)Wood, Li, Shaddick, and Augustin]{WoodBig}
S.~N. Wood, Z.~Li, G.~Shaddick, and N.~H. Augustin.
\newblock Generalized {A}dditive {M}odels for {G}igadata: {M}odeling the
  {U}.{K}. {B}lack {S}moke {N}etwork {D}aily {D}ata.
\newblock \emph{Journal of the American Statistical Association}, 112\penalty0
  (519):\penalty0 1199--1210, 2017.

\bibitem[Yee and Wild(1996)]{YeeWild1996}
T.~W. Yee and C.~J. Wild.
\newblock Vector {G}eneralized {A}dditive {M}odels.
\newblock \emph{Journal of the Royal Statistical Society, Series B},
  58\penalty0 (3):\penalty0 481--493, 1996.
\newblock ISSN 00359246.
\newblock URL \url{http://www.jstor.org/stable/2345888}.

\end{thebibliography}

\end{document}